\documentclass[preprint,12pt]{elsarticle}

\usepackage{graphicx}
\usepackage{hyperref}
\usepackage[utf8]{inputenc}
\usepackage{threeparttable}
\usepackage{longtable}

\usepackage{amsmath}
\usepackage{amsfonts}
\usepackage{siunitx}
\usepackage{soul}
\usepackage{xcolor}
\newcommand{\mycomment}[1]{}
\usepackage{booktabs}
\usepackage[ruled, vlined]{algorithm2e}

\usepackage{adjustbox}
\usepackage{float}
\usepackage{multirow}
\usepackage{subcaption}
\usepackage{makecell}
\usepackage[top=1in, bottom=1in, left=1in, right=1in]{geometry}
\usepackage{microtype}

\graphicspath{{images/}}
\journal{Biosystems Engineering}

\biboptions{authoryear}

\DeclareUnicodeCharacter{2009}{\,}
\hypersetup{
  pdftitle={LettuceVisSim: A Simulator That Generates Lettuce Image Time-series for Vision-Based Reinforcement Learning},
  pdfauthor={Ziye Zhu, Bert van t Ooster, Congcong Sun, Eldert van Henten, Sjoerd Boersma},
  pdfkeywords={Controlled Environment Agriculture, Vision-Based Reinforcement Learning, Synthetic Data Generation, Machine Vision, Crop Growth Model},
}

\begin{document}
    \begin{frontmatter}
        \title{LettuceVisSim: A Simulator That Generates Lettuce Image Time-series for Vision-Based Reinforcement Learning}

        \author{Ziye Zhu\fnref{myfootnote1}\corref{cor1}}
        \author{Bert van 't Ooster\fnref{myfootnote1}}
        \author{Congcong Sun\fnref{myfootnote1}}
        \author{Eldert van Henten\fnref{myfootnote1}}
        \author{Sjoerd Boersma\fnref{myfootnote2}}
        \fntext[myfootnote1]{Agricultural Biosystems Engineering Group, Wageningen University \& Research, The Netherlands}
        \fntext[myfootnote2]{Biometris, Wageningen University \& Research, The Netherlands}
        \cortext[cor1]{Corresponding email: ziye.zhu@wur.nl}

        \begin{abstract}
        \small
            Vision-based reinforcement learning holds strong potential for decision-making in controlled environment agriculture (CEA). However, its development is hindered by the scarcity of labelled crop images. To address this gap, LettuceVisSim, a lettuce growth simulator that generates labelled time series of crop images, was developed and validated. The simulator contains a process-based model (PBM) for shoot dry weight dynamics, a canopy layout algorithm for deriving canopy layout representations from shoot dry weight, and a Unity rendering engine for image generation. Five findings support the simulator. First, the PBM reproduced shoot dry weight under dynamic plant-density management with $\mathrm{R}^{2}=0.84$. Second, a piecewise cubic regression mapped shoot dry weight to potential projected area with $\mathrm{R}^{2}=0.94$. Third, the canopy layout representation was validated using 12 experimental datasets each having different dynamic environmental and spacing conditions. It reproduced the ground coverage ratio dynamics observed in measured images, achieving $\mathrm{R}^{2}=0.84$ when driven by measured shoot dry weight and $\mathrm{R}^{2}=0.40$ (0.76 excluding one outlier) when driven by PBM-simulated values. Fourth, the Unity rendering engine converted canopy layout representations into RGB and segmentation images at less than 10~ms. Fifth, a demonstration showed that a lighting-control policy can be learned and applied by observing only crop images that were generated with LettuceVisSim, providing a proof of concept of vision-based reinforcement learning in CEA using LettuceVisSim.
        \end{abstract}

    \end{frontmatter}

    \section{Introduction}
    \label{sec:introduction}

    Controlled environment agriculture (CEA) refers to food production systems in which crop growth takes place under artificially regulated indoor environmental conditions, it includes the greenhouse and plant factory production systems \citep{kozai_plant_2019,chen_review_2025}. By decoupling crop production from external weather variability, CEA offers a promising approach to address the growing global demand for food under increasing constraints on land and water availability, as well as rising climate change risks \citep{hossain_agricultural_2020,guo_agricultural_2016}. To fully realise this potential, a key challenge lies in optimising crop growth, which requires the appropriate setting of climate control setpoints governing the growth environment \citep{henten_greenhouse_1994,bakker_greenhouse_1995}.

    Many control methods have been studied for this task. Optimal control has been used for lettuce in both greenhouses \citep{henten_greenhouse_1994,xu_adaptive_2018} and plant factories \citep{xu_optimal_2021} and model predictive control to greenhouse tomato production \citep{hu_renewable_2022,xu_rule-based_2025}. More recently, reinforcement learning (RL), a data-driven, learning-based control method \citep{sutton_reinforcement_2018}, has emerged as a promising new option, with reports for greenhouse tomato \citep{an_simulator-based_2021,laatum_greenlight-gym_2024} and lettuce \citep{morcego_reinforcement_2023,mallick_reinforcement_2025} production.

    In these control studies, the crop state is usually the canopy biomass (dry weight, $kg \cdot m^{-2}$), and assumed to be measured. In practice, however, dry weight is not frequently measured by destructive sampling. Such measurements are not accurate, and they make crop data scarce. This creates a clear gap between what these controllers assume and what is achievable in CEA, which limits the knowledge of their true potential \citep{gautron_reinforcement_2022}.

    Vision-based RL \citep{mnih_human-level_2015} holds potential to mitigate this gap, as it learns control directly from crop images rather than from biomass measurements. It is promising for two reasons. First, crop image acquisition is non-destructive and second, a time series of a biomass proxy can be estimated from them \citep{ibaraki_plant_2014,bumgarner_digital_2012,concepcion_ii_lettuce_2020,du_greenhouse-based_2021}. Its main limitation, however, is data scarcity. Vision-based RL needs large amounts of labelled training data. One simple grasping task alone requires about 580,000 trajectories \citep{kalashnikov_qt-opt_2018}. The datasets at such scale are generally not available in agriculture \citep{parr_multimodal_2021}.

    A common approach to mitigate this scarcity is generating data with simulators. Such simulators have been widely adopted in robotics \citep{zhao_sim--real_2020}, drones \citep{song_flightmare_2021} and process control systems \citep{li_reinforcement_2021}. To support development of vision-based RL in CEA, a simulator should generate time-series of crop images that are labelled with corresponding quantitative crop traits such as shoot dry weight, representing the crop state transition under dynamic environmental and management conditions and do so efficiently.

    However, no existing model meets this need. Existing approaches fall into two types. Process-based models (PBM) simulate biomass from physiological mechanisms \citep{di_paola_overview_2016}. These models cannot generate images. Another type of existing models are the Functional--structural plant models (FSPM) can render plant morphology \citep{buck-sorlin_functional-structural_2013}, but these models are often static or computationally expensive \citep{louarn_two_2020}. Beyond crop growth models, generation models such as diffusion model can generate highly realistic images \citep{croitoru_diffusion_2022,tan_generative_2025,zangana_diffusion_2025}, but they are slow at inference and provide no explicit link between pixels and traits such as shoot dry weight. Hence, no current method can generate labelled time-series crop images at the throughput that vision-based RL requires.
    
    In this study, we follow the hierarchical design proven in drone simulation \citep{song_flightmare_2021}, where a PBM computes the system state. This state is then mapped to a visual representation, after which a game engine renders it into images. This design was applied to lettuce growth and present LettuceVisSim, an integrated lettuce simulator that generates time-series top-view canopy images labelled with shoot dry weight, as a function of temperature, light, CO$_{2}$, and plant spacing. LettuceVisSim consists of three main modules:
    
    \begin{enumerate}
        \item A lettuce PBM simulating shoot dry weight in response
            to dynamic environmental conditions and spacing management.

        \item A canopy layout algorithm that converts shoot dry weight into a canopy layout
            representation.

        \item A Unity rendering engine that generates images from canopy layout
            representation.
    \end{enumerate}

    For the PBM, the well-established Van Henten lettuce model \citep{van_henten_validation_1994} was adopted. It takes, among other, plant density as an input, but was developed and validated under a fixed density. CEA, however, routinely reduces density as the canopy expands, so its validity under dynamic spacing needs to be examined.

    The canopy layout algorithm consists of two steps. First, an empirical model maps shoot dry weight to the potential projected area (PPA), the top-view area of an isolated plant. Second, a uniform-grid construction combines PPA and plant density into a canopy layout that approximates the top-view canopy. From this layout, the ground coverage ratio (GCR) could be computed in the same way as for measured images.

    The Unity rendering engine then turns the canopy layout into top-view RGB and segmentation images, where the efficiency of image generation is the primary concern.

    This study validated LettuceVisSim as a data generator for vision-based RL through five research questions:

    \begin{enumerate}
        \item Does the PBM, originally developed under a fixed plant density, remain valid for predicting shoot dry weight dynamics under dynamic plant-density management?

        \item How can shoot dry weight be accurately mapped to PPA across growth stages?

        \item Can the proposed canopy layout representation reproduce GCR dynamics consistent with those derived from measured images under dynamic environmental and spacing conditions?
        
        \item How efficiently can a Unity rendering engine generate images in terms of visual fidelity and computational performance?

        \item Can images generated by the validated simulator, in principle, support vision-based RL training?
    \end{enumerate}
    
    \section{Material and methods}

    \subsection{Simulator overview}
\begin{figure}[htbp]
        \centering
        \includegraphics[width=1\linewidth]{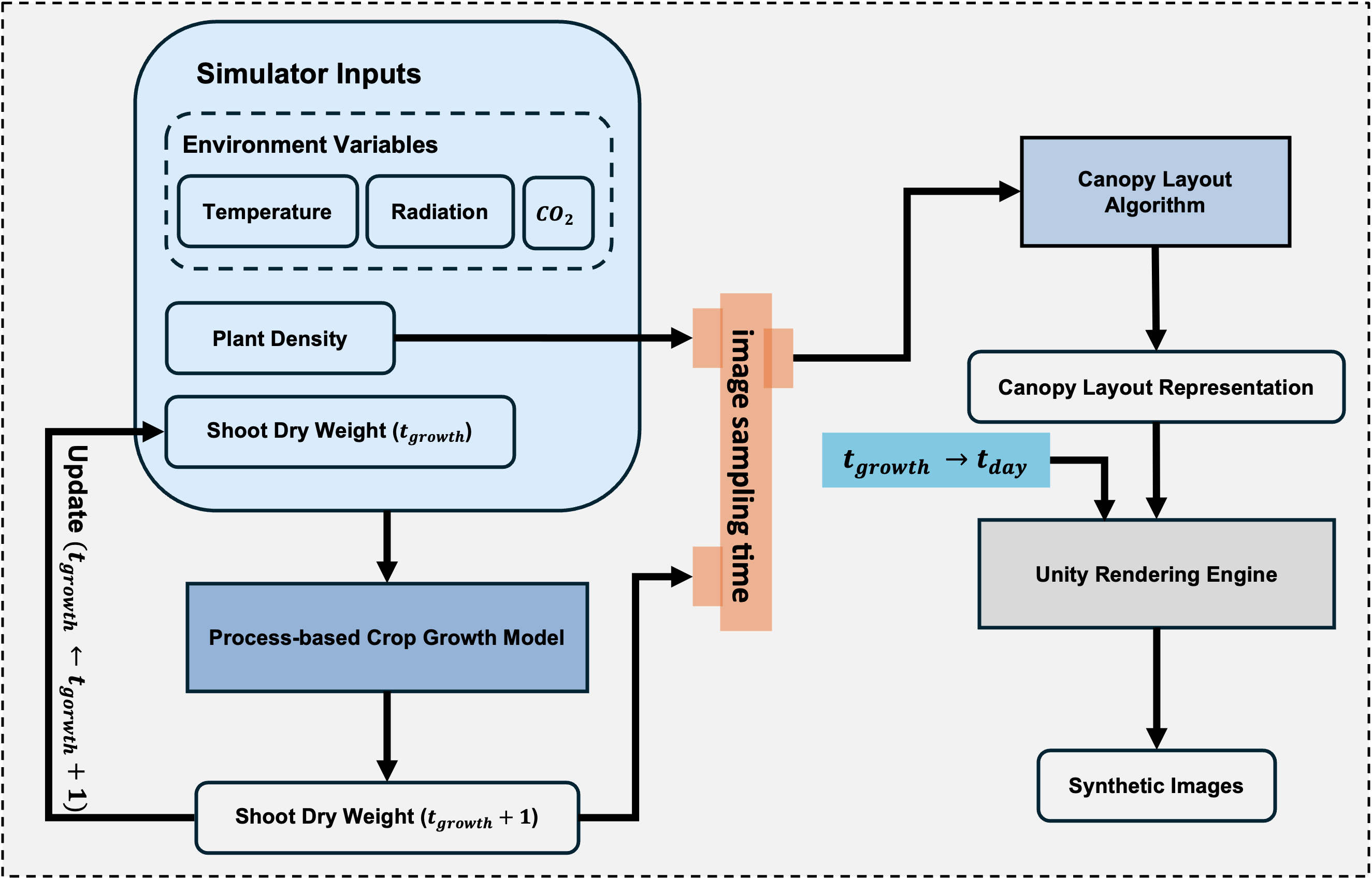}
        \caption{Schematic diagram of LettuceVisSim. Rectangular boxes represent the three modules of the simulator: the Process-Based Model (PBM), the canopy layout algorithm, and the Unity rendering engine. Rounded boxes represent the output of each module: simulated shoot dry weight~$\hat{w}$, canopy layout representation~$\mathcal{C}$, and synthetic RGB and segmentation images, respectively. At each image sampling time, the canopy layout algorithm and the Unity rendering engine are called sequentially to map the current shoot dry weight to synthetic images.}
        \label{fig:simulator_overview}
    \end{figure}

    LettuceVisSim is a modular lettuce growth simulator consisting of three modules, as shown in Fig.~\ref{fig:simulator_overview}: a \textbf{PBM} (Section~\ref{lab:pbm}) that simulates biomass accumulation in response to environmental variables and spacing management, a canopy layout algorithm (Section~\ref{lab:canopy-layout-algorithm}) that converts shoot dry weight per plant into a canopy layout representation, and a Unity rendering engine (Section~\ref{lab:unity-rendering-engine}) that generates synthetic images based on the canopy layout representation.

    In this simulator, shoot dry weight accumulation is simulated by the PBM at a temporal resolution $\Delta t_{\mathrm{growth}}=5$~min. The image sampling interval is independently configurable, provided it is no shorter than $\Delta t_{\mathrm{growth}}$. At each image sampling time, the canopy layout algorithm converts the simulated shoot dry weight and plant density into a canopy layout representation, which is then passed together with the elapsed cultivation day $t_{day}$, defined as the integer number of days since transplanting, to the Unity rendering engine to generate top-view RGB images and corresponding segmentation masks.

    The PBM used in this study has been validated and widely applied in many control studies \citep{boersma_nonlinear_2022,svensen_chance-constrained_2024,xu_optimal_2020} and is therefore only briefly described in the following section. The canopy layout algorithm and the Unity rendering engine are newly proposed in this study and are described in detail in the following sections.

    \subsubsection{Process based model}
    \label{lab:pbm}

    This section presents a brief overview of the PBM used in this study; a detailed description is provided in Van Henten \citep{van_henten_validation_1994}. In the original PBM, lettuce growth is formulated in continuous time as an ordinary differential equation (ODE). In this study, a discrete-time representation was obtained by numerically integrating the ODE with a fixed time step $\Delta t_{\mathrm{growth}}$. Let $\hat{w}$ denote simulated shoot dry weight (g per plant). The PBM can be written as:
    
    \begin{equation}
        \frac{\mathrm{d}\hat{w}_{t}}{\mathrm{d}t}=g\!\left(\hat{w}_{t},\,T_{t},\,I_{t},\,CO_{2,t},\,pd_{t},p\right),
    \end{equation}
    
    \noindent where $T$, $I$, $CO_{2}$, $pd$ denote air temperature ($^{\circ}$C), incident radiation ($\mu\mathrm{mol}\cdot \mathrm{m}^{-2}\cdot \mathrm{s}^{-1}$),
    carbon dioxide concentration concentration (ppm) and plant density (plants $\cdot \mathrm{m}^{-2}$), respectively, and $p$ denotes model parameters. The state update used in the simulator was obtained with forward Euler integration:
    
    \begin{equation}
        \hat{w}_{t+1}=\underbrace{\hat{w}_{t}+\Delta t_{\mathrm{growth}}\; g\!\left(\hat{w}_{t},\,T_{t},\,I_{t}
        ,\,CO_{2,t},\,pd_{t}, \,p\right)}_{f\!\left(\hat{w}_{t},\,T_{t},\,I_{t},\,CO_{2,t},\,pd_{t},\,p\right)}, \label{eq:pbm_discrete_map}
    \end{equation}
    \noindent where $\hat{w}_{t}\approx \hat{w}(t_{0}+t\,\Delta t_{\mathrm{growth}})$ and the inputs were assumed constant over each interval of length $\Delta t_{\mathrm{growth}}$. The detailed PBM equations and parameter definitions are provided in \ref{appendix:pbm}.

    \subsubsection{Canopy layout algorithm}
    \label{lab:canopy-layout-algorithm}
    
    The canopy layout algorithm converts the shoot dry weight input $w^{\ast}$ and plant density $pd_t$ into a canopy layout representation at each image sampling time step, where $w^{\ast}=w$ when measured shoot dry weight is used and $w^{\ast}=\hat{w}$ when simulated shoot dry weight is used. This canopy layout representation describes a $1~\mathrm{m}^{2}$ top view of the canopy in a simplified form, as a set of circles in which each circle approximates the top-view projection of an individual plant. Each circle is parameterised by its centre coordinates and its radius. The canopy layout algorithm comprises a shoot dry weight--PPA mapping step and a uniform-grid construction step, as described next.

    The \textit{shoot dry weight--PPA mapping step} defines an empirical regression model, denoted as $f_{\mathrm{PPA}}: w^{\ast} \mapsto \mathrm{PPA}$, to convert shoot dry weight per plant $w^{\ast}$ into the corresponding PPA. Empirical regression models have been widely used to predict biomass from image-based traits such as PPA \citep{li_describing_2022, baek_non-destructive_2025}. Common formulations include linear, polynomial, logistic, and hyperbolic regression models, which can be used in an inverse manner to map shoot dry weight to PPA. Because such formulations apply a single function over the entire growth period, growth-stage-dependent changes in the relation between PPA and shoot dry weight may not be captured adequately. Therefore, in addition to these so called one-phase formulations, a two-phase formulation was considered, in which the growth period is divided into two growth phases separated by a transition point. Both one-phase and two-phase empirical regression models were treated as candidate implementations of $f_{\mathrm{PPA}}$. Their mathematical formulations, parameter estimation, and comparative evaluation for model selection are described later in Experiment~2.

    The \textit{uniform-grid construction step} uses $f_{\mathrm{PPA}}$ to convert shoot dry weight $w^{\ast}$ and plant density $pd_t$ into a canopy layout representation $\mathcal{C}$, which approximates the top-view lettuce canopy within the camera field of view (FoV). This representation corresponds to the top-view canopy captured by a fixed overhead camera mounted above the centre of the cultivation area, with a monitored area of $1~\mathrm{m}^{2}$. This construction relies on three simplifying assumptions: (1) plants are distributed uniformly within the cultivation area; (2) boundary effects are taken into account by including partially visible plants along the edges of the image; and (3) canopy closure is modelled as geometric overlap of PPA, neglecting physical deformation or architectural adaptation caused by mechanical contact between neighboring plants. Besides $w^{\ast}$ and $pd_t$, the FoV dimensions $(L \times W)$ are also taken as input.

    The \textit{uniform-grid construction step} proceeds in four stages: grid initialisation, grid expansion for boundary handling, radius computation and FoV-intersection filtering, and canopy layout construction. In grid initialisation stage, the FoV aspect ratio $\alpha=L/W$ and plant density $pd_t$ are used to determine the grid dimensions, from which a regular set of plant centres $P_{\text{opt}}$ is generated within the FoV. In grid expansion for boundary handling stage, this grid is expanded by integer shifts along both spatial directions, yielding an extended candidate set $P_{\text{ext}}$. In radius computation and FoV-intersection filtering stage, the projection radius is computed as $r=\sqrt{f_{\mathrm{PPA}}(w^{\ast})/\pi}$, and only those centres whose circular projections intersect the FoV are retained, resulting in $P_{\text{final}}$. In the canopy layout construction stage, the canopy layout representation $\mathcal{C}$ is constructed as the set of circles centred at $P_{\text{final}}$ with common radius $r$. Together, these four stages define the canopy layout algorithm, which is summarised in Algorithm~\ref{alg:ugcc}.

    \begin{figure}[H]
        \centering
        \resizebox{\textwidth}{!}{%
        \begin{minipage}{\textwidth}
            \begin{algorithm}
                [H] \footnotesize
                \caption{Canopy layout algorithm}
                \label{alg:ugcc} \KwIn{$w^{\ast}$: shoot dry weight per plant; $pd_t$: plant density;\\ \ \ \ \ \ \ \ \  $L \times W$: size of field of view (FoV), corresponding to a top-view area of 1 $\mathrm{m}^{2}$}
                \KwOut{Canopy layout representation $\mathcal{C}$ as a set of circular canopy projections in two-dimensional space}

                \tcp{1. Grid initialisation}
                $\alpha \gets L/W$ \; $col \gets \left\lceil \sqrt{ pd_t\cdot
                \alpha}\right\rceil$ \; $row \gets \left\lceil \frac{pd_t}{col}\right\rceil$

                $dx \gets \frac{L}{col+1}, \quad dy \gets \frac{W}{row+1}$

                $P_{\text{opt}}\gets \{(i \cdot dx,\ j \cdot dy)\ |\ i=1,\dots,co
                l;\ j=1,\dots,row\}$

                \tcp{2. Grid expansion for boundary handling} $x_{\min}, x_{\max}
                \gets \min(x), \max(x)$ in $P_{\text{opt}}$ \; $y_{\min}, y_{\max}
                \gets \min(y), \max(y)$ in $P_{\text{opt}}$ \;

                $s_{\text{left}}\gets \left\lceil \tfrac{x_{\min}}{dx}\right\rceil
                ,\; s_{\text{right}}\gets \left\lceil \tfrac{L-x_{\max}}{dx}\right
                \rceil,\; s_{\text{bottom}}\gets \left\lceil \tfrac{y_{\min}}{dy}
                \right\rceil,\; s_{\text{top}}\gets \left\lceil \tfrac{W-y_{\max}}
                {dy}\right\rceil$

                $X_{\text{shift}}\gets \{-s_{\text{left}}, \dots, s_{\text{right}}
                \} \cdot dx$ \; $Y_{\text{shift}}\gets \{-s_{\text{bottom}}, \dots
                , s_{\text{top}}\} \cdot dy$ \;

                $P_{\text{ext}}\gets \{(x+\delta_{x},\ y+\delta_{y})\ |\ (x,y)\in
                P_{\text{opt}},\ \delta_{x}\in X_{\text{shift}},\ \delta_{y}\in Y
                _{\text{shift}}\}$

                \tcp{3. Radius computation and FoV-intersection filtering}
                $PPA \gets f_{\mathrm{PPA}}(w^{\ast})$ \; $r \gets \sqrt{PPA / \pi}$ \tcp*[l]{Shoot dry weight--PPA mapping step}
                $P_{\text{valid}}\gets \{(x,y)\in P_{\text{ext}}\mid -r \le x \le
                L+r,\ -r \le y \le W +r\}$
                \; $P_{\text{final}}\gets \text{unique}(P_{\text{valid}})$ \;

                \tcp{4. Canopy layout construction}

                $\mathcal{C}\gets \{ \text{Circle}(x,y,r)\ |\ (x,y) \in P_{\text{final}}
                \}$ \;

                \Return $\mathcal{C}$
            \end{algorithm}
        \end{minipage}
        }
    \end{figure}

    \subsubsection{Unity rendering engine}
    \label{lab:unity-rendering-engine}

\begin{figure}[H]
        \centering
        \includegraphics[width=1\linewidth]{
            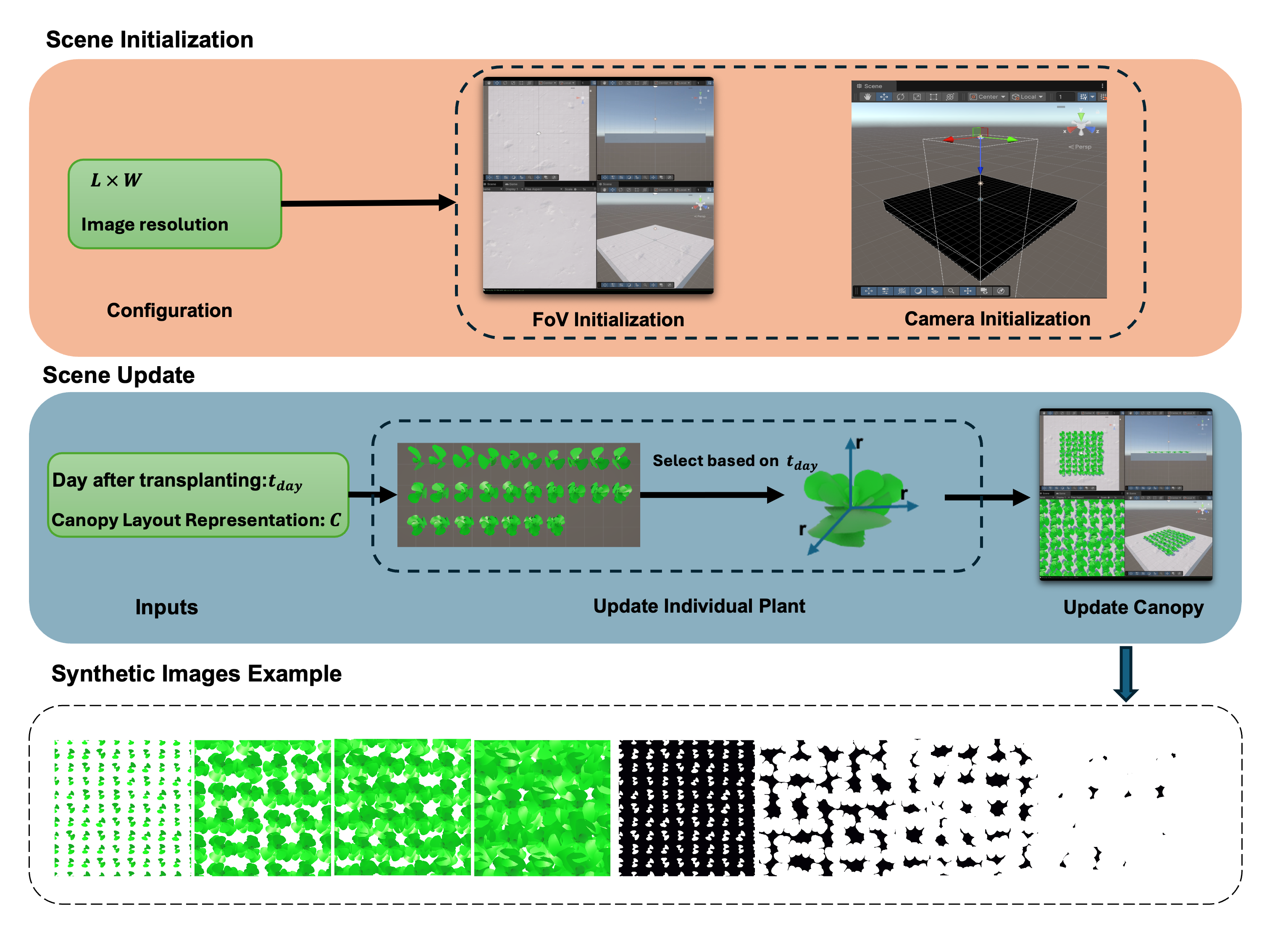
        }
        \caption{Overview of two steps in the Unity rendering engine for synthetic image generation. Scene initialisation uses predefined configurations to initialise the scene prior to simulation. Scene updates reconstruct the canopy appearance based on the canopy layout representation, from which synthetic RGB and segmentation images are rendered with a virtual camera.}
        \label{fig:engine_overview}
    \end{figure}

    The Unity rendering engine generates synthetic top-view RGB images together with corresponding binary segmentation masks, by capturing the reconstructed canopy appearance using a virtual camera. The canopy appearance is reconstructed from the canopy layout representation. As illustrated in Fig.~\ref{fig:engine_overview}, the rendering engine operates in two steps: scene initialisation and scene update.

    Scene initialisation is performed prior to simulation, using the initial configuration provided as inputs. The FoV of $1~\mathrm{m}^{2}$ is initialised as a rectangular domain of $L \times W$. The virtual camera is configured with a user-specified image resolution. A top-down orthographic projection is used to generate the top-view images, which preserves a consistent pixel-to-metric scaling across the FoV and eliminates perspective distortion.

    The scene update is performed at each image sampling time step, after the canopy layout representation has been generated by the canopy layout algorithm. The previously generated canopy layout representation and the simulated day $t_{day}$ are taken as input in order to construct the canopy appearance. The update process consists of two stages: individual plant updating and overall canopy reconstruction.

    For individual plant updating, lettuce morphological development is approximated by a simple rule: each seedling has three leaves at transplanting, the leaf number increases by one per day, and is capped at 34 leaves (i.e., when $t_{day}\geq 30$). Based on this rule, a library of 31 prefabricated lettuce models was constructed in Blender, each representing a specific leaf number from 3 to 34. Leaves were sampled from real leaf images that covered all growth stages, and each model was constructed by stacking leaves with their tips arranged in a spiral around the centroid. For a given $t_{day}$, the corresponding prefab is selected and scaled uniformly in all three dimensions by means of using the radius parameter $r$ in $\mathcal{C}$. When desired, a small random in-plane rotation and a scale perturbation can be applied per plant instance to introduce limited plant-level variability.

    The canopy appearance is then constructed by placing the selected and scaled prefabs within the FoV according to the final plant positions $P_{\text{final}}$ in $\mathcal{C}$. Each prefab is positioned by aligning the centre of its base with the corresponding coordinates. Finally, the updated scene is rendered by the virtual camera, which outputs the corresponding top-view RGB image and binary segmentation mask.

        \subsection{Data acquisition}
    \label{lab:data-acquisition}
    This section introduces the datasets used for developing and validating LettuceVisSim. The Data description section details two publicly available datasets, their experimental setups, and the types of data collected. The Data Selection and Processing section outlines the steps taken to ensure consistency in cultivar selection, image quality, and preprocessing for extracting relevant features for development and validation.

    \subsubsection{Data description}
    \label{lab:data-description}
    Publicly available datasets from the 3rd Autonomous Greenhouse Challenge were used in this study. The datasets were collected during lettuce cultivation experiments conducted in Bleiswijk, The Netherlands, in 2021 and 2022. Specifically, two datasets with different experimental setups were used, and they are briefly described below. Detailed protocols are provided in the original publications.
    
    \paragraph{Dataset (i)}  This dataset, collected in 2021, comprises top-view RGB-D images of individual lettuce plants from four cultivars, including \textit{cv.~Lugano}, spanning a range of developmental stages from seedling to harvest. Each plant image is paired with its corresponding shoot dry weight $w_i$ and additional phenotypic traits \citep{hemming_3rd_2021}.

    \paragraph{Dataset (ii)} This dataset was collected in February--March and May--June 2022, during two cultivation cycles of \textit{cv.~Lugano}, each under 6 distinct cultivation strategies, yielding 12 strategies in total. This dataset comprises hourly top-view RGB-D time-series images of lettuce canopy over an area of 1~$\text{m}^{2}$. In addition to image data in dataset (ii), synchronised environmental and management records were collected, together with weekly destructive dry weight measurements \citep{petropoulou_lettuce_2023}. Specifically, the multimodal time-series dataset across 12 cultivation strategies, including canopy images, climate variables, spacing management records, and shoot dry weight measurements, was used in this study. A summary of the environmental conditions and strategy abbreviations, which refer to the participating team name and cultivation cycle, is provided in Table~\ref{tab:env_summary}.

    \begin{table}[H]
    \centering
    \caption{Summary of realised indoor climate variables for all teams across two cultivation cycles. Min and Max denote the minimum and maximum recorded values, and SD denotes the standard deviation}
    \label{tab:env_summary}
    \renewcommand{\arraystretch}{1.2}

    \begin{adjustbox}{max width=\textwidth}
        \begin{tabular}{clcccccccccc}
            \toprule
            \textbf{Cultivation cycle} 
                & \textbf{Team} 
                & \multicolumn{3}{c}{\textbf{Air remperature (°C)}} 
                & \multicolumn{3}{c}\textbf{Indoor radiation (W m$^{-2}$)}
                & \multicolumn{3}{c}\textbf{Carbon dioxide concentration (ppm)}
                & \textbf{Abbreviation} \\
            
            \cmidrule(lr){3-5}
            \cmidrule(lr){6-8}
            \cmidrule(lr){9-11}
            
                & 
                & \textbf{Min} 
                & \textbf{Mean $\pm$ SD} 
                & \textbf{Max} 
                & \textbf{Min} 
                & \textbf{Mean $\pm$ SD} 
                & \textbf{Max} 
                & \textbf{Min} 
                & \textbf{Mean $\pm$ SD} 
                & \textbf{Max} 
                & \\
            \midrule

            \multirow{6}{*}{2 February to March 2022}
                & cva
                & 12.60 & 20.05$\pm$3.67 & 30.20
                & 0     & 19.24$\pm$26.47 & 114.62
                & 408.00 & 489.00$\pm$117.35 & 974.00
                & cva1 \\
            & koala
                & 11.20 & 19.63$\pm$3.36 & 28.40
                & 0     & 40.76$\pm$24.92 & 96.79
                & 399.00 & 811.01$\pm$233.54 & 1279.00
                & koala1 \\
            & monday-lettuce
                & 15.30 & 21.88$\pm$3.26 & 29.60
                & 0     & 33.92$\pm$34.68 & 114.84
                & 361.00 & 704.97$\pm$218.26 & 1568.00
                & monday-lettuce1 \\
            & digital-cucumbers
                & 15.50 & 19.78$\pm$2.03 & 29.10
                & 0     & 26.43$\pm$25.32 & 100.92
                & 501.00 & 919.74$\pm$95.04 & 1780.00
                & digital-cucumbers1 \\
            & veggie-might
                & 10.10 & 19.80$\pm$3.63 & 30.60
                & 0     & 36.49$\pm$25.37 & 107.66
                & 415.00 & 667.48$\pm$119.05 & 906.00
                & veggie-might1 \\
            & reference
                & 12.00 & 17.87$\pm$2.90 & 29.20
                & 0     & 22.61$\pm$18.03 & 90.92
                & 393.00 & 596.26$\pm$75.73 & 1067.00
                & reference1 \\
            
            \midrule

            \multirow{6}{*}{4 May to June 2022}
                & cva
                & 12.90 & 19.63$\pm$4.05 & 31.60
                & 0     & 31.16$\pm$44.38 & 383.45
                & 383.00 & 547.45$\pm$158.26 & 1338.00
                & cva2 \\
            & koala
                & 11.70 & 20.17$\pm$3.57 & 30.80
                & 0     & 32.99$\pm$46.69 & 373.23
                & 360.00 & 463.00$\pm$83.74 & 1123.00
                & koala2 \\
            & monday-lettuce
                & 13.90 & 21.67$\pm$4.00 & 32.40
                & 0     & 27.20$\pm$32.04 & 359.09
                & 330.00 & 488.68$\pm$142.22 & 1025.00
                & monday-lettuce2 \\
            & digital-cucumbers
                & 17.00 & 22.48$\pm$3.25 & 36.70
                & 0     & 26.55$\pm$47.18 & 382.15
                & 427.00 & 757.95$\pm$179.11 & 1329.00
                & digital-cucumbers2 \\
            & veggie-might
                & 14.10 & 20.49$\pm$3.54 & 31.20
                & 0     & 44.51$\pm$69.04 & 372.14
                & 385.00 & 486.11$\pm$71.07 & 1085.00
                & veggie-might2 \\
            & reference
                & 13.90 & 21.23$\pm$3.84 & 32.40
                & 0     & 27.95$\pm$33.91 & 359.09
                & 330.00 & 495.55$\pm$137.72 & 1012.00
                & reference2 \\
            
            \bottomrule
        \end{tabular}
    \end{adjustbox}
\end{table}

    \subsubsection{Data selection and processing}
    \label{lab:data-selection}

    Selection was applied to both Dataset (i) and Dataset (ii) for two purposes. First, the selection ensured that both datasets finally contained growth measurements of the same lettuce cultivar, thereby supporting cultivar-consistent growth modelling and cross-dataset validation. Second, the selection guaranteed that crop pixels were clearly captured in the images, allowing extraction of image-based features.

\begin{figure}[H]
        \centering
        \includegraphics[width=1\linewidth]{
            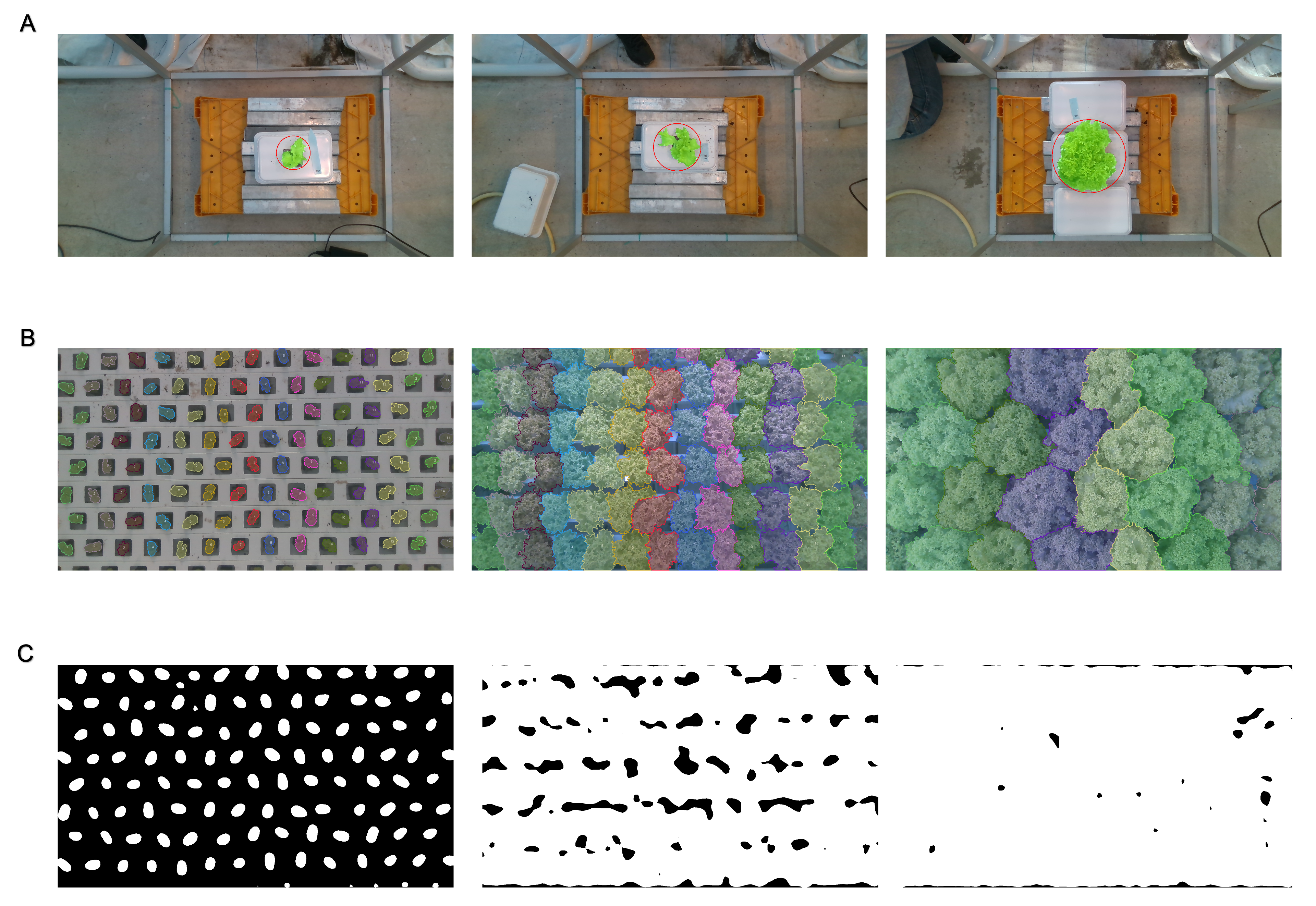}
        \caption{Examples of image annotation and predicted masks across different datasets. (A) Annotation of lettuce plants and visualisation of their minimum enclosing circles in Dataset (i); (B) annotation of lettuce plants within the canopy of Dataset (ii); (C) corresponding predicted masks generated by the trained DeepLabv3+ model from the input images shown in panel (B).}
        \label{fig:image_annotation}
    \end{figure}

    \paragraph{Dataset (i)}
    Only images and corresponding measurements of \textit{cv.~Lugano} were used. After selection, images were processed to extract PPA. As exemplified in Fig.~\ref{fig:image_annotation}(A), the lettuce plants in each image were first segmented manually from the background using the Segment Anything Model \citep{kirillov2023segany}. The minimum enclosing circle of each segmented lettuce region was then computed, together with the pixel count inside the circle. The physical area of the circle was obtained from the measured plant diameter, and the PPA of image $j$, denoted ($\text{PPA}_j$) was estimated via the resulting pixel-to-area mapping. This process yielded 96 pairs of shoot dry weight and PPA measurements, defined as the dataset:

    \begin{equation}
    \label{eq:dataset_1}
            \mathcal{D}^{\text{i}}=\{(w_j,\,{\text{PPA}}_j)\}_{j=1}^{N=96}
    \end{equation}
    
\begin{figure}[H]
        \centering
        \includegraphics[width=1\linewidth]{
            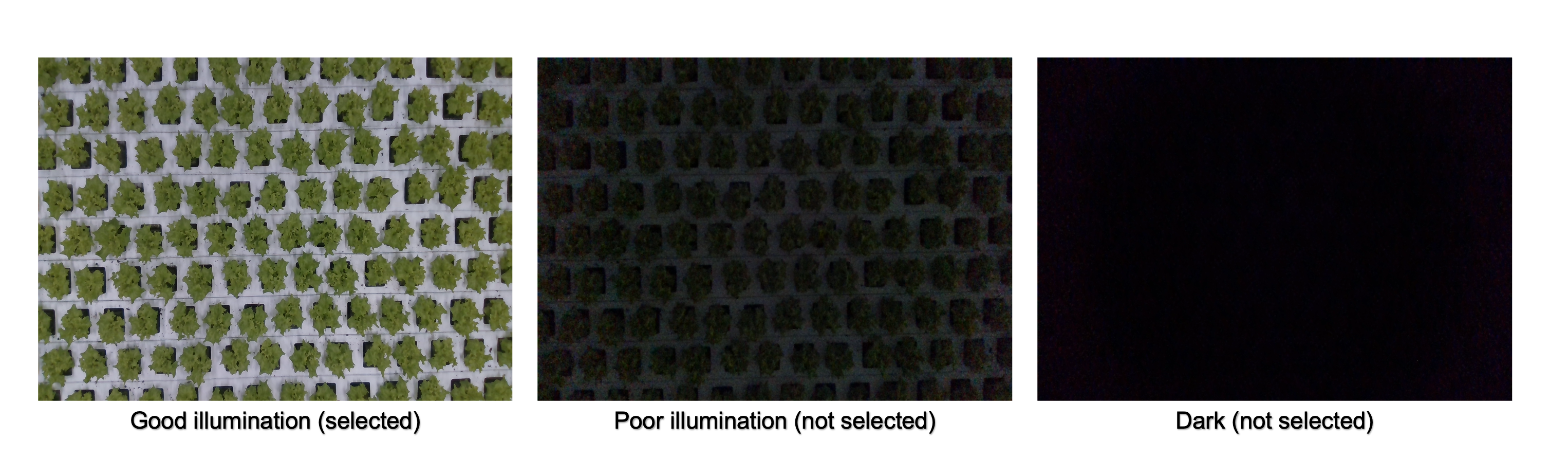
        }
        \caption{Illustrative examples of image screening based on illumination conditions in Dataset (ii). From left to right: a representative image with good canopy visibility (indoor photosynthetically active radiation above 5~$\mu\text{mol m}^{-2}\text{ s}^{-1}$), an image captured when illumination was lower than the threshold, and an image captured under purely dark conditions.}
        \label{fig:illumination_screen_example}
    \end{figure}

    \paragraph{Dataset (ii)}
    Images were first screened based on illumination conditions. Only images captured during periods in which the indoor photosynthetically active radiation exceeded 5~$\mu\text{mol m}^{-2}\text{ s}^{-1}$ were included, ensuring that canopy was clearly visible in the images, as shown in Fig.~\ref{fig:illumination_screen_example}. After applying this threshold, approximately 40\% (4942 images) of the original images were retained.

    After selection, images were processed to estimate GCR, following the protocol of \citet{petropoulou_lettuce_2023}. Images were first sampled from all cultivation strategies at fixed 7-day intervals, starting from the first day of each cultivation cycle. For each sampling day, two images were selected at 09:00 and 16:00 to capture canopy appearance under stable morning and afternoon lighting conditions prior to annotation. This sampling procedure yielded 131 canopy images for annotation, in which all lettuce plants with visually distinguishable boundaries were manually labelled, as exemplified in Fig.~\ref{fig:image_annotation}(B).
    
    After annotation, the images were split randomly into training, validation, and test sets in a 6:2:2 ratio and used to train a DeepLabv3+ model \citep{chen_encoder-decoder_2018} for semantic segmentation, following the same approach as \citet{petropoulou_lettuce_2023}. A mean intersection over union (mIoU, Eq.~\eqref{eq:miou}) of 93.5\% was achieved on the test set (\ref{appendix:deeplabv3}). Compared to \citet{petropoulou_lettuce_2023}, who used 23 annotated images for training and 12 for validation, this study used substantially more annotated samples, providing a stronger basis for model training. Once trained, the model was applied to all remaining selected images in Dataset~(ii) to generate the corresponding segmentation masks. For each generated mask, as exemplified in Fig.~\ref{fig:image_annotation}(C), GCR was computed as the ratio of the number of pixels classified as canopy to the total number of pixels, which yielded a GCR time series.

    Beyond GCR estimation, these annotated images were also used to quantify the number of lettuce plants visible within the camera field of view. Because the original plant density records $pd_t$ were available only at a daily resolution, hourly measured time-series images were additionally inspected to determine density changes based on temporal canopy development, which enables simulation at shorter time intervals. The relation between $pd_t$ and the visible plant counts derived from the annotated images is given in \ref{pd-and-plants-count}.
    
    In addition to these image-derived quantities, the weekly destructive shoot dry weight measurements from Dataset (ii) are formally defined as:
    \begin{equation}
    \label{eq:dataset_2}
        \mathcal{D}^{\text{ii}} = \{\,w_{s,t} \mid s \in \mathcal{S},\; t \in \mathcal{T}_{s}\,\},
    \end{equation}
    \noindent where $\mathcal{S}$ denotes the set of 12 cultivation strategies and $\mathcal{T}_{s}$ denotes the set of destructive sampling time points for strategy $s$. Hence, $w_{s,t}$ is the shoot dry weight of cultivation strategy $s$ at time $t$.

        \subsection{Experiment description}
    \label{lab:experiments}
    
    This section describes the experimental procedures used to develop, validate, and evaluate the LettuceVisSim. Experiment~1 examined whether the PBM, originally developed under a fixed plant density, remained valid for predicting shoot dry weight dynamics under dynamic plant-density management. Experiment~2 compared regression models for mapping shoot dry weight to PPA and selected the best-performing model for the canopy layout algorithm. Experiment~3 validated the canopy layout representation by comparing simulated GCR dynamics with those estimated from measured images under varying environmental and spacing conditions, considering both measured and PBM-simulated shoot dry weight as inputs. Experiment~4 evaluated the Unity rendering engine in terms of output diversity and computational efficiency for synthetic image generation. In Experiment~5, a proof of concept of vision-based RL with LettuceVisSim was provided.

    \subsubsection{Experiment 1: validation of the process-based model under dynamic plant-density management}
    \label{lab:exp1}

    In Experiment~1, the PBM (Section~\ref{lab:pbm}) for shoot dry weight prediction under dynamic plant-density management was validated with the measured shoot dry weight $\mathcal{D}^{\text{ii}}$ (Eq.~\eqref{eq:dataset_2}) and the synchronised environmental and plant-density records of the 12 strategies in $\mathcal{S}$ (Table~\ref{tab:env_summary}). 
    
    A six-fold leave-strategy-out cross-validation was used to evaluate the PBM on strategies not seen during calibration \citep{roberts_cross-validation_2017}. First, the 12 strategies in $\mathcal{S}$ were split randomly into six folds of two strategies each. Second, the following steps were performed in each run $k$ ($k=1,\dots,6$):
    
    \begin{enumerate}
        \item One fold served as the held-out test set $\mathcal{S}_{\mathrm{test}}^{k}$ and the other five as the training set $\mathcal{S}_{\mathrm{train}}^{k}$.
        \item In the calibration step, the optimisation in~\eqref{eq:pbm_calibration_objective} was solved. 
        \item In the evaluation step, the trained model was tested on $\mathcal{S}_{\mathrm{test}}^{k}$. 
    \end{enumerate}
    
    In the calibration step, based on the sensitivity analysis of the PBM \citep{van_henten_sensitivity_1994}, only the five parameters most influential for biomass accumulation, $p_{\mathrm{cal}}=(c_b,\ c_e,\ c_{\mathrm{gr,max}},\ c_K,\ c_{\mathrm{lar}})^{T}$, were calibrated, with the remaining parameters held fixed at their original values (\ref{appendix:pbm}). A parameter set $p_{\mathrm{cal}}$ was estimated by running the PBM (Eq.~\eqref{eq:pbm_discrete_map}) forward over the training strategies and minimising the squared error between the simulated shoot dry weight $\hat{w}_{s,t}(p_{\mathrm{cal}})$ and the measured shoot dry weight $w_{s,t}$,
    
    \begin{equation}
        p_{\mathrm{cal},k}^{\ast}
        =
        \arg\min_{p_{\mathrm{cal}}}
        \sum_{s \in \mathcal{S}_{\mathrm{train}}^{k}}
        \sum_{t \in \mathcal{T}_{s}}
        \Big(\hat{w}_{s,t}(p_{\mathrm{cal}})-w_{s,t}\Big)^{2},
        \label{eq:pbm_calibration_objective}
    \end{equation}

    where $w_{s,t}\in\mathcal{D}^{\text{ii}}$ and $\mathcal{T}_s$ are the destructive sampling times (Eq.~\eqref{eq:dataset_2}). The optimisation was carried out with the \texttt{scipy.optimize} module of SciPy.

    In the evaluation step, the calibrated PBM was run forward on each of the two strategies $s\in\mathcal{S}_{\mathrm{test}}^{k}$ to produce two shoot dry weight trajectories, and the six runs together yielded 12 test trajectories. 
    
    Performance was computed by comparing simulated shoot dry weights with the corresponding measurements. It was quantified by the coefficient of determination ($\mathrm{R}^{2}$), measuring how well the simulation captured the observed variation in shoot dry weight, and the relative root mean squared error (RRMSE), preferred over absolute error because shoot dry weight spans a wide range across growth stages and strategies. Both are defined in \ref{appendix:metrics} and reused in the subsequent experiments.

    These metrics were computed at both the fold and strategy levels. At the fold level, the two held-out strategies in each $\mathcal{S}_{\mathrm{test}}^{k}$ were pooled to obtain one $\mathrm{R}^{2}$ and one RRMSE, as is commonly done in crop model studies \citep{droutsas_integration_2022,hussain_simulation_2026}. At the strategy level, the two metrics were instead computed on each of the 12 test trajectories separately. This strategy-wise reporting keeps poorly fitted strategies from being hidden by pooling and offers a more complete view of model performance \citep{bates_cross-validation_2024}.
    
    The same 12 test trajectories were subsequently used as the PBM-driven shoot dry weight inputs to the canopy layout algorithm in Experiment~3.

    \subsubsection{Experiment 2: model selection for the shoot dry weight--PPA mapping}
    \label{lab:exp2}
    
    Experiment~2 compared regression models and selected the one used to serve as $f_{\mathrm{PPA}}$ in the simulator. Specifically, this comparison considered a set of one-phase regression models, and a two-phase piecewise cubic regression was included as the proposed approach. The model formulations are given below.

    \textbf{One-phase models:} Shoot dry weight is mapped to PPA by a single function over the entire shoot dry weight range,
    \begin{align}
        \text{Linear:}~~     y(w^{\ast}) &= \beta_{\mathrm{lin}}\,w^{\ast} + \alpha_{\mathrm{lin}}, \\
        \theta_{\mathrm{lin}} &= \begin{pmatrix} \beta_{\mathrm{lin}} & \alpha_{\mathrm{lin}} \end{pmatrix}^T. \nonumber \\
        \text{Quadratic:}~~  y(w^{\ast}) &= \beta_{2}^{\mathrm{quad}}(w^{\ast})^{2} + \beta_{1}^{\mathrm{quad}}w^{\ast} + \alpha^{\mathrm{quad}}, \\
        \theta_{\mathrm{quad}} &= \begin{pmatrix} \beta_{2}^{\mathrm{quad}} & \beta_{1}^{\mathrm{quad}} & \alpha^{\mathrm{quad}} \end{pmatrix}^T. \nonumber \\
        \text{Cubic:}~~      y(w^{\ast}) &= \beta_{3}^{\mathrm{cub}}(w^{\ast})^{3} + \beta_{2}^{\mathrm{cub}}(w^{\ast})^{2} + \beta_{1}^{\mathrm{cub}}w^{\ast} + \alpha^{\mathrm{cub}}, \\
        \theta_{\mathrm{cub}} &= \begin{pmatrix} \beta_{3}^{\mathrm{cub}} & \beta_{2}^{\mathrm{cub}} & \beta_{1}^{\mathrm{cub}} & \alpha^{\mathrm{cub}} \end{pmatrix}^T. \nonumber \\
        \text{Quartic:}~~    y(w^{\ast}) &= \beta_{4}^{\mathrm{qr}}(w^{\ast})^{4} + \beta_{3}^{\mathrm{qr}}(w^{\ast})^{3} + \beta_{2}^{\mathrm{qr}}(w^{\ast})^{2} + \beta_{1}^{\mathrm{qr}}w^{\ast} + \alpha^{\mathrm{qr}}, \\
        \theta_{\mathrm{qr}} &= \begin{pmatrix} \beta_{4}^{\mathrm{qr}} & \beta_{3}^{\mathrm{qr}} & \beta_{2}^{\mathrm{qr}} & \beta_{1}^{\mathrm{qr}} & \alpha^{\mathrm{qr}} \end{pmatrix}^T. \nonumber \\
        \text{Logistic:}~~   y(w^{\ast}) &= \dfrac{L_{\mathrm{log}}}{1 + e^{-\,k_{\mathrm{log}}\,(w^{\ast} - w_{0,\mathrm{log}})}}, \\
        \theta_{\mathrm{log}} &= \begin{pmatrix} L_{\mathrm{log}} & k_{\mathrm{log}} & w_{0,\mathrm{log}} \end{pmatrix}^T. \nonumber \\
        \text{Hyperbolic:}~~ y(w^{\ast}) &= \dfrac{a_{\mathrm{hyp}}}{\,w^{\ast} - w_{0,\mathrm{hyp}}\,} + b_{\mathrm{hyp}}, \\
        \theta_{\mathrm{hyp}} &= \begin{pmatrix} a_{\mathrm{hyp}} & w_{0,\mathrm{hyp}} & b_{\mathrm{hyp}} \end{pmatrix}^T, \nonumber
    \end{align}

    \textbf{Two-phase model:} Because the relation between shoot dry weight and PPA may vary, a piecewise cubic regression is proposed. It is composed of two cubic polynomial segments, $f_{1}$ and $f_2$, joined at a breakpoint $w_{b}$, with continuity in both the function value and the first derivative,
    \begin{equation}
    \begin{aligned}
    \text{Piecewise:}~~ y(w^{\ast}) &=
    \begin{cases}
    f_1 = a_0^1+ a_1^1w^{\ast} + a_2^1(w^{\ast})^{2}+ a_3^1(w^{\ast})^{3}, & w^{\ast} < w_{b}, \\
    f_2 =b_0^2+ b_1^2w^{\ast} + b_2^2(w^{\ast})^{2}+ b_3^2(w^{\ast})^{3}, & w^{\ast} \geq w_{b},
    \end{cases} \\[0.5ex]
    \text{subject to}\quad
    & f_{1}(w_{b}) = f_{2}(w_{b}), \\
    & f_{1}'(w_{b}) = f_{2}'(w_{b}).
    \end{aligned}
    \label{eq:piecewise_regression}
    \end{equation}
    
    \noindent with $\theta_{\mathrm{pw}} = \begin{pmatrix} a_0^1 & a_1^1 & a_2^1 & a_3^1 & b_0^2 & b_1^2 & b_2^2 & b_3^2 \end{pmatrix}^T$ the parameter vector that has been optimised. This approach is intended to capture the early and late growth stages, separately. The prime symbol ($'$) denotes the derivative with respect to $w^{\ast}$. Note again that $w^{\ast}=w$ when measured shoot dry weight is used and $w^{\ast}=\hat{w}$ when shoot dry weight simulated by PBM, (Eq.~\eqref{eq:pbm_discrete_map}) is used

    Experiment~2 was conducted on paired shoot dry weight--PPA dataset $\mathcal{D}^{\text{i}}$ (Eq.~\eqref{eq:dataset_1}) with five-fold cross-validation. This procedure was used to compare candidate model structures and breakpoint settings. The set $\mathcal{D}^{\text{i}}$ was sorted in ascending order of shoot dry weight and partitioned into five folds, ensuring that the full biomass range was represented in each fold. In the $k^{\text{th}}$ cross-validation run, four folds formed the training set $\mathcal{D}^{\text{i}}_{\mathrm{train},k}$, while the remaining fold formed the held-out test set $\mathcal{D}^{\text{i}}_{\mathrm{test},k}$.
    
    For a given regression model $y(\cdot;\theta_\mathrm{m})$ for $\mathrm{m}=\{\text{lin},\text{quad},\text{cub}, \text{qr},\text{log},\text{hyp},\text{pw}\}$, this procedure was repeated for all five folds, such that each fold served once as the test set. The model parameters in each run were estimated by using \texttt{scipy.optimize} module in SciPy to minimise the sum of squared residuals on the training set,

    \begin{equation}
    \hat{\theta}_\mathrm{m}^{(k)}
    =
    \arg\min_{\theta_\mathrm{m}}
    \sum_{j\in\mathcal{D}^{\text{i}}_{\mathrm{train},k}}
    \Big(
    {\text{PPA}}_j-y(w_j;\theta_\mathrm{m})
    \Big)^2,
    \label{eq:ppa_foldwise_fitting}
    \end{equation}
    
    \noindent where $\hat{\theta}_\mathrm{m}^{(k)}$ denotes the estimated parameters of the model in the $k^{\text{th}}$ run. For all one-phase models ($\mathrm{m} \ne \text{pw}$), parameters were estimated using the full training set. For the two-phase model ($\mathrm{m} = \text{pw}$), the parameters of each segment were estimated on the corresponding sub-range of the training set, the parameters of $f_1$ on training samples with $w < w_{b}$, and the parameters of $f_2$ on training samples with $w \ge w_{b}$, subject to the continuity conditions in Eq.~\eqref{eq:piecewise_regression}. The fitted model's predictive performance was then calculated from $\mathcal{D}^{\text{i}}_{\mathrm{test},k}$ using each $\hat{\theta}_\mathrm{m}^{(k)}$ for $k=1,\ldots,5$.

    Predictive performance was quantified with the same $\mathrm{R}^{2}$ and RRMSE metrics introduced in Experiment~1, here applied to PPA and summarised over the five folds.

    The proposed two-phase model was first compared with the one-phase models with a fixed breakpoint $w_{b}=0.7~\mathrm{g}$. Performance was evaluated over the entire shoot dry weight range, as well as separately for samples with $w < w_{b}$ and $w\ge w_{b}$ in order to examine stage-specific fitting behaviour.

    The sensitivity of model performance to breakpoint selection was then assessed by varying $w_{b}$ from $0.3$ to $1.1~\mathrm{g}$ in increments of $0.2~\mathrm{g}$. For each candidate breakpoint, the two-phase model was fitted and evaluated with the same cross-validation procedure.

    Based on cross-validation results, the model with the best overall performance was selected. The selected model was then refitted on the complete dataset $\mathcal{D}^{\text{i}}$ with the same least-squares fitting procedure as described above, i.e.,
    \begin{equation}
        \hat{\theta}_{\mathrm{m}}^{*}=\arg\min_{\theta_m}\sum_{j\in\mathcal{D}^{\text{i}}}\Big({\text{PPA}}_j-y(w_j;\theta_m)\Big)^2. 
    \end{equation}
    Here, $\hat{\theta}_{\mathrm{m}}^{*}$ is subsequently used in Experiment~3 to construct the canopy layout representation, which is evaluated on Dataset~(ii). Dataset~(ii) was collected independently, in a different experiment from Dataset~(i).

    \subsubsection{Experiment 3: validation of the canopy layout representation}
    \label{lab:exp3}

    Experiment~3 validated whether the canopy layout representation could reproduce GCR dynamics consistent with those derived from measured images under dynamic environmental and spacing conditions. The representation was driven by two sources of shoot dry weight: (i) the measured shoot dry weight, which isolates the validity of the representation itself under accurate inputs, and (ii) the PBM-simulated shoot dry weight from Experiment~1, which reflects the downstream GCR agreement obtained when the representation is driven by model-predicted rather than measured inputs.

    The simulated GCR was derived from the canopy layout representation with a common representation-to-GCR conversion procedure in both cases. The shoot dry weight input $w^{\ast}$ and the corresponding plant density $pd_t$ were used as inputs to the canopy layout algorithm to generate a canopy layout representation, which was then converted into a binary mask at the same image resolution as the measured images ($1920 \times 1080$), using the pixel-to-metric scaling of the field of view. This ensured consistency with the GCR estimation method applied to measured images. Pixels were marked as covered if their squared Euclidean distance to the nearest plant centre was less than or equal to the squared canopy radius $r$, and the GCR was calculated from the resulting binary mask. Agreement between simulated and image-derived GCR was quantified using $\mathrm{R}^{2}$ and RRMSE.

    For the validation driven by measured shoot dry weight, the measured shoot dry weight $w$ and the corresponding plant density were used as $w^{\ast}$ and $pd_t$. Because shoot dry weight was obtained by destructive sampling, this comparison was performed only on the days for which destructive measurements were available, and the simulated GCR was compared against the daily averaged image-derived GCR on those days. The daily reference GCR was obtained by averaging the image-derived GCR values from all observations on the same day, with the standard deviation used to characterise within-day variability.

    For the validation driven by PBM-simulated shoot dry weight, the test-strategy dry weight trajectories $\hat{w}_{t}$ from the six-fold leave-strategy-out cross-validation in Experiment~1, together with the corresponding plant density records $pd_{t}$, were used as $w^{\ast}$ and $pd_t$ to generate time series of canopy layout representations, from which GCR time series were derived. The resulting GCR time series were compared with the hourly image-derived GCR at matched time points.

    \subsubsection{Experiment 4: evaluation of the Unity rendering engine}
    \label{lab:exp4}

    This experiment evaluated the proposed Unity rendering engine in terms of output diversity under different rendering configurations and computational efficiency.

    The evaluations were conducted under a simulation scenario considering a 30-day cultivation period using real climate and plant spacing records from strategy \textit{cva1}. To demonstrate the ability of the engine to generate diverse visual outputs, identical simulation states were rendered under different configurations, including varying lighting levels, image resolutions, and randomised plant rotations using the selected prefabs. 

    Computational efficiency was quantified by the average computation time per frame, defined as the total wall-clock time divided by the number of generated frames, and assessed in two steps. First, the cost of individual stages was analysed by incrementing the execution scope: the PBM only, the PBM plus the canopy layout algorithm, and the full pipeline. For this analysis, the canopy layout representation was converted into a binary mask at an image resolution of $256 \times 256$~pixels with three image sampling intervals (5~min, hourly and daily), while rendered images were generated with a square field of view ($L = W = 1$~m) and an image resolution of $256 \times 256$ with two image sampling intervals (hourly and daily). Second, the full pipeline was evaluated under varying image sampling intervals (hourly and daily) and image resolutions ($84\times84$ to $1920\times 1080$). For each configuration, five independent runs were conducted to account for runtime variability.

    \subsubsection{Experiment 5: proof of concept of vision-based reinforcement learning with LettuceVisSim}
    \label{lab:exp5}

    Experiment~5 provided a proof of concept of the simulator for vision-based RL in a vertical farming control problem, in which crop growth is controlled by optimising the light based on crop image observations. The set-up is described below.

    \paragraph{Control task} A light-intensity control task over a fixed cultivation horizon of $H=30$ days is considered. The control objective is to steer the terminal shoot dry weight $\hat{w}_{H}$ into a harvest band of $[15,17]~\mathrm{g}\cdot\mathrm{plant}^{-1}$ while keeping the biomass trajectory non-decreasing throughout the cycle. Shoot dry weight accumulation is simulated by the PBM (Section~\ref{lab:pbm}), which updates the per-plant shoot dry weight $\hat{w}_{d}$ on day $d$ in response to the environmental inputs and a daily lighting setpoint. At the start of each day $d\in\{0,\dots,H-1\}$, the vision-based RL agent observes a crop image and, based on it, calculates a lighting setpoint $a_{d}\in[0,360]~\mu\mathrm{mol}\cdot\mathrm{m}^{-2}\cdot\mathrm{s}^{-1}$, expressed as a photosynthetic photon flux density (PPFD). This is then applied as incident radiation during the 18~h photoperiod and set to zero otherwise. Throughout the cultivation, air temperature is set to $22~^{\circ}$C during the light period and $20~^{\circ}$C during the dark period. The CO$_{2}$ concentration is held constant at $800$~ppm, and plant density is fixed at $18~\mathrm{plants}\cdot\mathrm{m}^{-2}$. 

    \paragraph{Image observation} At each day, the vision-based RL agent received a top-view canopy RGB image of $84\times84$ resolution as its sole observation and computed the lighting decision at a daily interval.

    \paragraph{Reward and PPO agent} The reward function $R(\hat{w}_{d})$ combines a dense per-day term with a sparse terminal term, which encourages the terminal shoot dry weight to land within the harvest band, while keeping the trajectory non-decreasing,

    \begin{equation}
        R(\hat{w}_{d}) =
        \begin{cases}
            \hat{w}_{d} - \hat{w}_{d-1}, & \hat{w}_{d} - \hat{w}_{d-1} \ge 0, \;\; d < H, \\
            -5,                          & \hat{w}_{d} - \hat{w}_{d-1} < 0, \;\; d < H, \\
            10,                          & \hat{w}_{H} \in [s_{\min}, s_{\max}], \;\; d = H, \\
            -10\,\left| \hat{w}_{H} - \dfrac{s_{\min}+s_{\max}}{2} \right|, & \text{otherwise}, \;\; d = H,
        \end{cases}
        \label{eq:rl_reward}
    \end{equation}

    where $[s_{\min},s_{\max}]=[15,17]~\mathrm{g}\cdot\mathrm{plant}^{-1}$ is the harvest band defined above. The first two terms return the one-day biomass gain and penalises any decrease in shoot dry weight, whereas the last two terms reward landing inside the harvest band and otherwise penalises the deviation from the band centre.

    Proximal Policy Optimization (PPO)~\citep{schulman_proximal_2017} was used as the vision-based RL algorithm in this demonstration, with an asymmetric actor--critic architecture: a convolutional encoder maps the image observation to the actor, which therefore depends on the image alone, while the critic additionally receives the shoot dry weight during training to stabilise value estimation. This keeps the deployed policy dependent on vision only. The network architecture and full PPO hyperparameters are detailed in \ref{appendix:rl}.

    \subsubsection{Experiment environment}
    All experiments were conducted on a Lenovo ThinkPad P15 Gen 2i laptop equipped with an Intel Xeon W-1885M CPU (12 cores, 3.2 GHz), 32 GB RAM, and an NVIDIA RTX A5000 laptop GPU (16 GB VRAM), running Pop!\_OS (64-bit). The simulation and data analysis were implemented in Python 3.10, using NumPy 1.26, OpenCV 4.9, PyTorch 2.0 and SciPy 1.15.3. The rendering engine was built in Unity (version 6000.0.48f1) and was executed in headless mode for automated simulation rendering.

    \section{Results}
    \label{sec:results}
    \subsection{Experiment 1}

    Table~\ref{tab:exp1_dw_cv} shows the fold-wise performance for PBM shoot dry weight prediction, for the training and test folds in each cross-validation run. Averaged over the six folds, the train agreement was $\mathrm{R}^{2}=0.873\pm0.031$ (RRMSE $37.6\pm4.9\%$) and the test agreement was $\mathrm{R}^{2}=0.836\pm0.131$ (RRMSE $38.6\pm17.1\%$), which indicates comparable agreement on the train and test strategies.

    Across folds, the train agreement stayed within a narrow range ($\mathrm{R}^{2}$ from $0.831$ to $0.921$), whereas the test agreement varied much more widely ($\mathrm{R}^{2}$ from $0.624$ to $0.955$). A similar pattern was observed for RRMSE: the train RRMSE remained within a narrow range ($29.5\%$ to $43.7\%$), whereas the test RRMSE spanned a much wider range ($20.5\%$ to $70.0\%$). The lowest test agreement occurred in Fold~5 (\textit{veggie-might1}, \textit{koala2}), with the lowest test $\mathrm{R}^{2}$ ($0.624$) and the highest test RRMSE ($70.0\%$) among all folds.

    \begin{table}[H]
        \centering
        \caption{Fold-wise performance of the six-fold leave-strategy-out cross-validation for shoot dry weight prediction by the process-based model, $\mathrm{R}^{2}$ denotes the coefficient of determination, RRMSE the relative root mean squared error and SD the standard deviation.}
        \label{tab:exp1_dw_cv}
        \begin{tabular}{clcccc}
            \toprule
            \multirow{2}{*}{Fold} & \multirow{2}{*}{Test strategies}
            & \multicolumn{2}{c}{Train} & \multicolumn{2}{c}{Test} \\
            \cmidrule(lr){3-4}\cmidrule(lr){5-6}
            & & $\mathrm{R}^{2}$ & RRMSE & $\mathrm{R}^{2}$ & RRMSE \\
            \midrule
            1 & \textit{cva2}, \textit{veggie-might2}            & 0.831 & 43.7\% & 0.955 & 20.5\% \\
            2 & \textit{koala1}, \textit{monday-lettuce2}        & 0.896 & 35.1\% & 0.683 & 52.2\% \\
            3 & \textit{digital-cucumbers1}, \textit{reference1} & 0.866 & 37.7\% & 0.911 & 32.2\% \\
            4 & \textit{monday-lettuce1}, \textit{reference2}    & 0.864 & 39.4\% & 0.925 & 27.6\% \\
            5 & \textit{veggie-might1}, \textit{koala2}          & 0.921 & 29.5\% & 0.624 & 70.0\% \\
            6 & \textit{cva1}, \textit{digital-cucumbers2}       & 0.858 & 40.3\% & 0.916 & 29.4\% \\
            \midrule
            \multicolumn{2}{c}{Mean $\pm$ SD} & 0.873 $\pm$ 0.031 & 37.6 $\pm$ 4.9\% & 0.836 $\pm$ 0.131 & 38.6 $\pm$ 17.1\% \\
            \bottomrule
        \end{tabular}
    \end{table}

\begin{figure}[H]
        \centering
        \includegraphics[width=\linewidth]{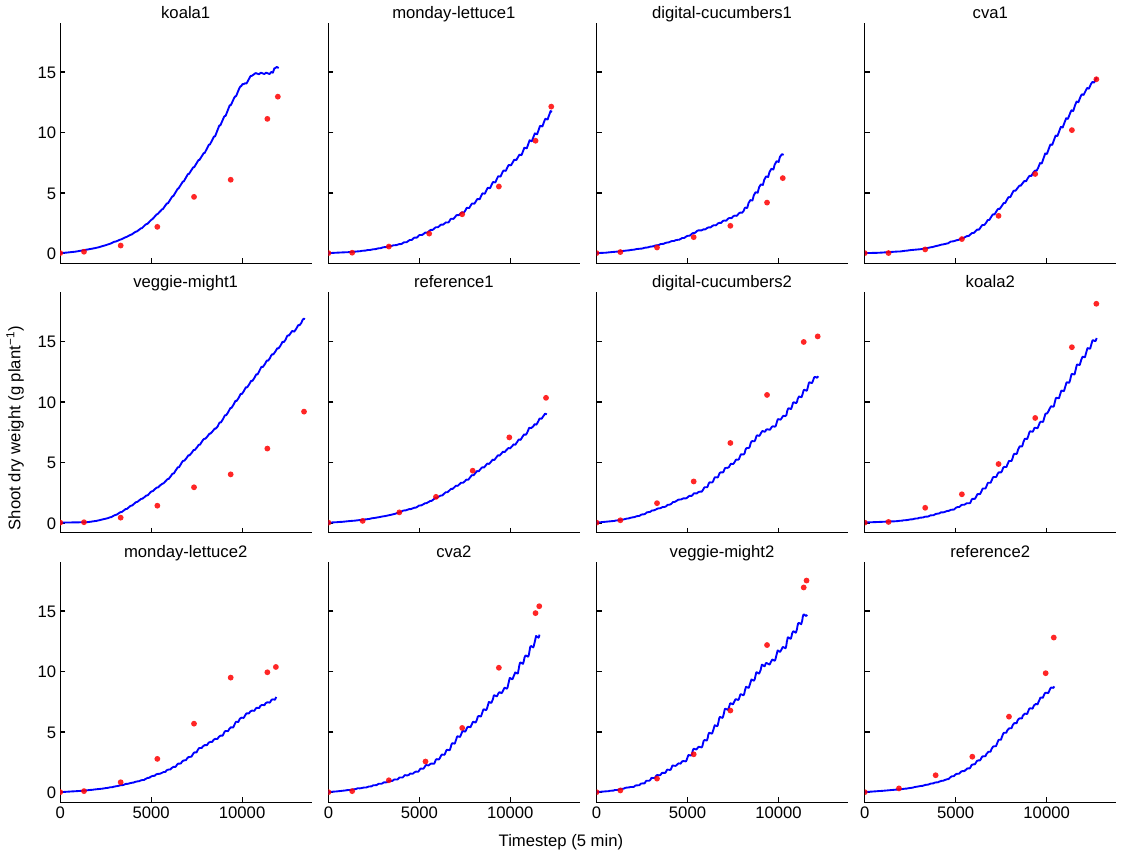}
        \caption{Shoot dry weight simulated by the process-based model (PBM) for the 12 cultivation strategies. For each strategy, the trajectory was generated with the parameter set calibrated on the cross-validation fold in which that strategy was held out, so every trajectory shown is an out-of-sample prediction. One timestep corresponds to 5~min. Blue lines denote the simulated shoot dry weight and red circles the destructively measured shoot dry weight. The corresponding per-strategy metrics are listed in Table~\ref{tab:F1}.}
        \label{fig:global_pbm_shoot_dry_weight}
    \end{figure}

    Whereas Table~\ref{tab:exp1_dw_cv} summarises agreement at the fold level, Fig.~\ref{fig:global_pbm_shoot_dry_weight} unpacks the same cross-validation into the 12 per-strategy test trajectories, each generated using the fold-specific parameters from the cross-validation run in which that strategy was held out. The simulated trajectories followed the general growth trend in most strategies, with fold-wise test $\mathrm{R}^{2}$ ranging from $0.624$ to $0.955$ (Table~\ref{tab:exp1_dw_cv}) and test $\mathrm{R}^{2}\ge0.91$ in four of the six folds. The largest deviation occurred in \textit{veggie-might1}, where the simulated shoot dry weight became much higher than the measured values at the late growth stage.

    \subsection{Experiment 2}

    Table~\ref{tab:regression_model_comparison} summarises the cross-validation performance of the proposed two-phase regression model with $w_{b}= 0.7 ~\mathrm{g}$ against the one-phase regression models. Overall, the two-phase model yielded best average performance across five folds ($\mathrm{R}^{2}=0.94$, RRMSE$=$22\%), though cubic and quartic models achieved near-identical overall metrics with overlapping fold-level standard deviations. Stage-wise results revealed that the clearest advantage of the two-phase model appeared in the early stage ($w<w_b$), where it achieved $\mathrm{R}^{2}=0.78$ and RRMSE$=$29\% while most one-phase models degraded substantially. In the late stage ($w\ge w_b$), model performance became closer, and the two-phase model remained competitive with the top one-phase models.

    \begin{table}[H]
        \centering
        \scriptsize
        \renewcommand{\arraystretch}{1.15}
        \caption{Comparison of one-phase and two-phase regression models for potential projected area (PPA) prediction across the overall dataset, the early stage ($w<0.7\,\mathrm{g}$) and the late stage ($w\geq0.7\,\mathrm{g}$). Values are reported as mean values over five cross-validation folds, with standard deviations given in parentheses.}
        \label{tab:regression_model_comparison}

        \makebox[\textwidth][c]{%
        \resizebox{\textwidth}{!}{%
        \begin{tabular}{llcccccc}
            \toprule
            \multirow{2}{*}{\textbf{Type}} & \multirow{2}{*}{\textbf{Model}}
            & \multicolumn{2}{c}{\textbf{Overall}}
            & \multicolumn{2}{c}{\textbf{Early stage}}
            & \multicolumn{2}{c}{\textbf{Late stage}} \\
            \cmidrule(lr){3-4} \cmidrule(lr){5-6} \cmidrule(lr){7-8}
            &
            & $\mathrm{R}^{2}$ & RRMSE
            & $\mathrm{R}^{2}$ & RRMSE
            & $\mathrm{R}^{2}$ & RRMSE \\
            \midrule
            \multirow{6}{*}{One-phase}
            & lin      & 0.91 (0.04) & 28 (4.08)\% & -0.27 (0.29) & 70 (8.41)\% & 0.82 (0.10) & 21 (3.71)\% \\
            & quad   & 0.92 (0.02) & 25 (2.93)\% & 0.46 (0.08) & 45 (4.60)\% & 0.84 (0.07) & 20 (2.64)\% \\
            & cub       & \textbf{0.94 (0.02)} & 23 (2.61)\% & 0.72 (0.05) & 33 (3.82)\% & \textbf{0.87 (0.05)} & 18 (2.36)\% \\
            & qr     & 0.94 (0.02) & 23 (2.73)\% & 0.74 (0.06) & 32 (3.89)\% & 0.86 (0.05) & 19 (2.47)\% \\
            & log    & 0.89 (0.03) & 31 (2.82)\% & -0.45 (0.30) & 74 (8.68)\% & 0.78 (0.07) & 24 (2.55)\% \\
            & hyp  & 0.93 (0.02) & 24 (2.85)\% & 0.67 (0.05) & 36 (3.51)\% & 0.85 (0.06) & 19 (2.58)\% \\
            Two-phase
            & pw   & \textbf{0.94 (0.02)} & \textbf{22 (2.90)\%} & \textbf{0.78 (0.04)} & \textbf{29 (3.02)\%} & \textbf{0.87 (0.05)} & \textbf{18 (2.57)\%} \\
            \bottomrule
        \end{tabular}%
        }}

        \vspace{0.3em}
        \makebox[\textwidth][c]{%
        \parbox{0.98\textwidth}{\footnotesize
        \textit{Note:} Mean value and standard deviations (in parentheses) are reported. Bold values indicate the best performance within each column, defined as the highest $\mathrm{R}^{2}$ and the lowest RRMSE. Model abbreviations are consistent with the $m$ notation defined in Section~\ref{lab:exp2}: lin (linear), quad (quadratic), cub (cubic), qr (quartic), log (logistic), hyp (hyperbolic) and pw (piecewise).}}
    \end{table}
    
\begin{figure}[H]
        \centering
        \includegraphics[width=1\linewidth]{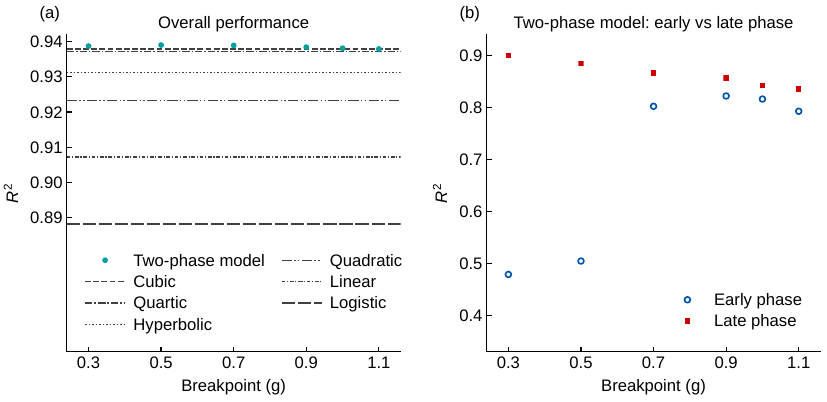}
        \caption{Left: Comparison of overall $\mathrm{R}^{2}$ between one-phase models and the two-phase model for describing the relationship between PPA and shoot dry weight, with the two-phase model evaluated across candidate breakpoints ($w_{b}=0.3$--$1.1\,\mathrm{g}$). Right: Changes in $\mathrm{R}^{2}$ for the early and late stages of the two-phase model across the same breakpoint range. All values represent the average performance over five-fold cross-validation.}
        \label{fig:breakpoint_sensitivity}
    \end{figure}

    Fig.~\ref{fig:breakpoint_sensitivity} illustrates the sensitivity of $\mathrm{R}^{2}$ to the breakpoint $w_{b}$ (from $0.3$ to $1.1~\mathrm{g}$). Although the overall $\mathrm{R}^{2}$ remained relatively insensitive to $w_{b}$, the highest overall $\mathrm{R}^{2}$ of $0.94$ was consistently achieved by the two-phase model across the evaluated breakpoint range, which exceeded all one-phase models (Fig.~\ref{fig:breakpoint_sensitivity}(a)). By contrast, the stage-specific trends responded differently to a change in breakpoint (Fig.~\ref{fig:breakpoint_sensitivity}(b)): the early-stage performance improved with larger $w_{b}$, stabilising around $w_{b}=0.7~\mathrm{g}$, whereas late-stage performance decreased monotonically as $w_{b}$ increased.

    On the basis of the results above, two-phase model was selected and $w_{b}= 0.7$~g was identified as a conservative breakpoint for subsequent analyses. This value corresponds to the smallest breakpoint at which early-stage performance stabilises, without compromising late-stage performance. The final $f_{\mathrm{PPA}}$ used in subsequent experiments takes the form,

    \begin{equation}
        f_{\mathrm{PPA}}(w^{\ast};\theta_{\mathrm{pw}})= 10^{-3}
        \begin{cases}
            1.444+13.444w^{\ast}-3.336(w^{\ast})^{2}-1.767(w^{\ast})^{3}, & w^{\ast}<0.7,    \\
            3.789+7.206w^{\ast}-0.592(w^{\ast})^{2}+0.024(w^{\ast})^{3},  & w^{\ast}\ge 0.7,
        \end{cases}
        \label{eq:fppa_final}
    \end{equation}
    
    \noindent in which $f_{\mathrm{PPA}}$ is expressed in m$^{2}$ and $w^{\ast}$ in g~plant$^{-1}$. The optimised (rounded) parameter vector is \\
    $\theta_{\mathrm{pw}} = 10^{-3}\times\begin{pmatrix} 1.444 & 13.444 & -3.336 & -1.767 & 3.789 & 7.206 & -0.592 & 0.024 \end{pmatrix}^T$. The shoot dry weight input, corresponding to either the measured shoot dry weight $w$ or the PBM-simulated shoot dry weight $\hat{w}$, as defined in Section~\ref{lab:canopy-layout-algorithm}.

    \subsection{Experiment 3}

\begin{figure}[H]
        \centering
        \includegraphics[width=1\linewidth]{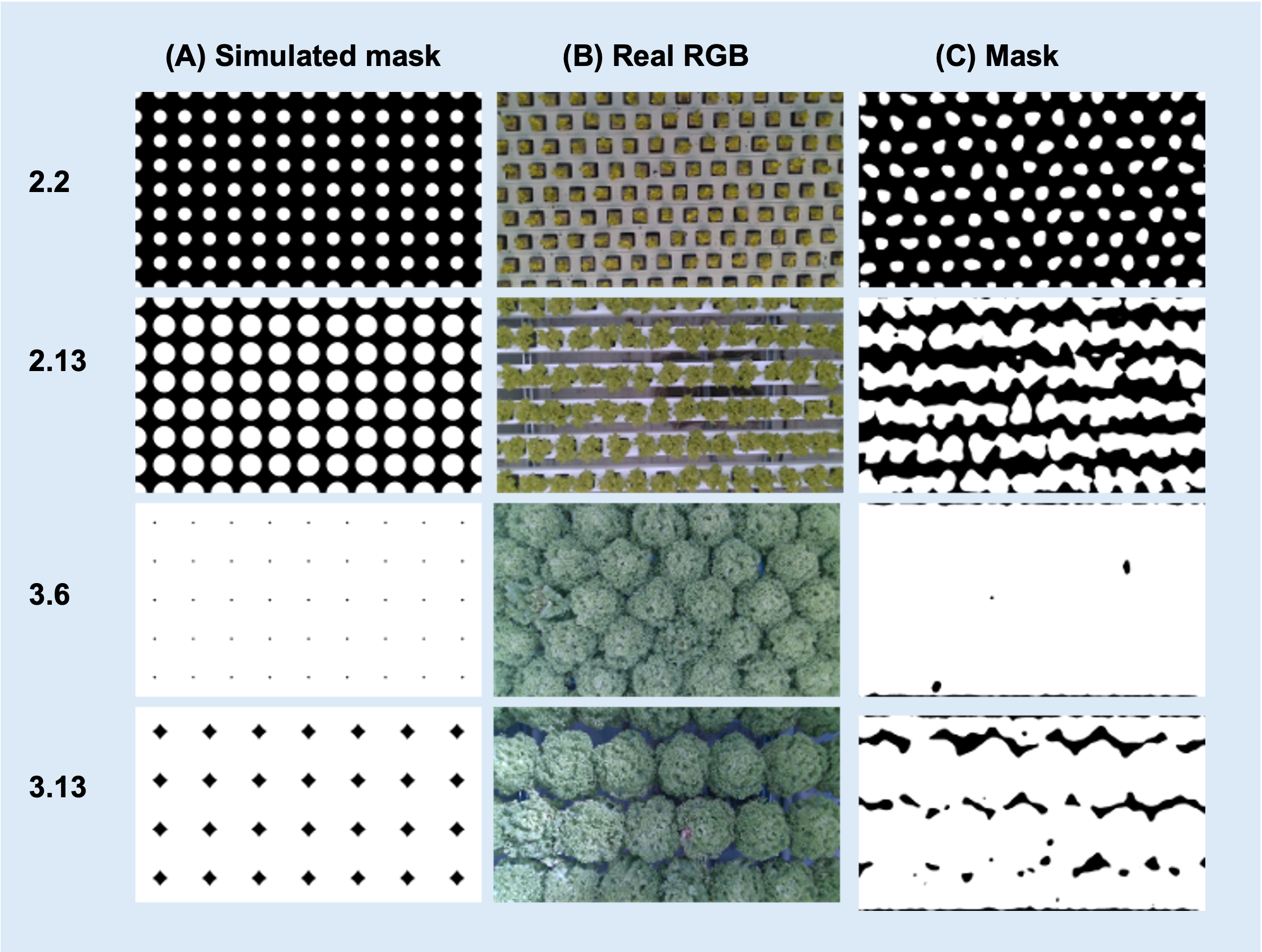}
        \caption{Visual comparison between binary masks generated from the canopy layout representation, with their RGB images and corresponding masks at different sampling dates from the \textit{cva1} strategy. The numbers on the left indicate the corresponding sampling dates. (A) Binary masks derived from the canopy layout representation, simulated using measured shoot dry weight on each sampling date. (B) Measured RGB canopy images acquired at 12:00 on the same dates, and (C) the corresponding binary masks segmented using the trained DeepLabv3+.}
        \label{fig:layout_mask_visualization}
    \end{figure}

    Fig.~\ref{fig:layout_mask_visualization} presents a visual comparison between canopy masks generated from canopy layout representations using measured shoot dry weight as input and canopy masks segmented from measured RGB images on the same day. At early growth stages, the simulated and measured masks showed similar canopy occupancy patterns. As cultivation progressed, structural differences became more visible: the simulated masks retained a more regular grid-based arrangement, whereas the measured masks showed more irregular uncovered areas and plant positions.

\begin{figure}[H]
        \centering
        \includegraphics[width=\linewidth]{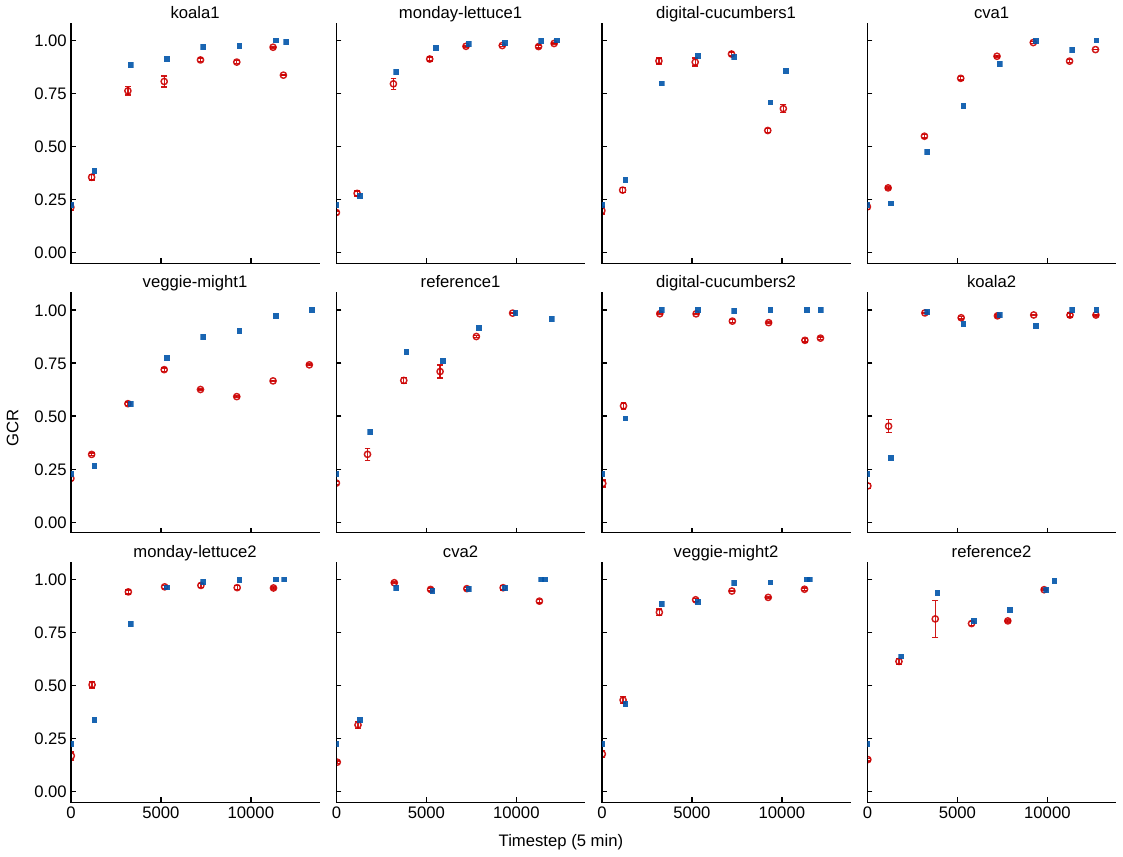}
        \caption{Ground coverage ratio (GCR) simulated using measured shoot dry weight across cultivation strategies. Simulated GCR (blue) is compared with the daily averaged image-derived GCR measurements (red), with error bars indicating the standard deviation across observations from the same day. The per-strategy metrics are listed in Table~\ref{tab:F2}.}
        \label{fig:GCR_under_measured_dry_weight}
    \end{figure}

    Fig.~\ref{fig:GCR_under_measured_dry_weight} presents the quantitative comparison between GCR simulated from measured shoot dry weight and GCR derived from same-day images across 12 strategies. Overall agreement was strong, with $\mathrm{R}^{2}=0.84$ and RRMSE ranging from 4.2\% to 36.3\%. Most strategies showed close agreement between simulated and observed daily GCR. The main exception was observed in strategy \textit{veggie-might1}, where a large deviation appeared after day 20, resulting in a negative $\mathrm{R}^{2}$ and the largest RRMSE. Excluding this strategy increased the overall performance to $\mathrm{R}^{2}=0.93$, with a mean RRMSE of 9.4\%.

\begin{figure}[H]
        \centering
        \includegraphics[width=\linewidth]{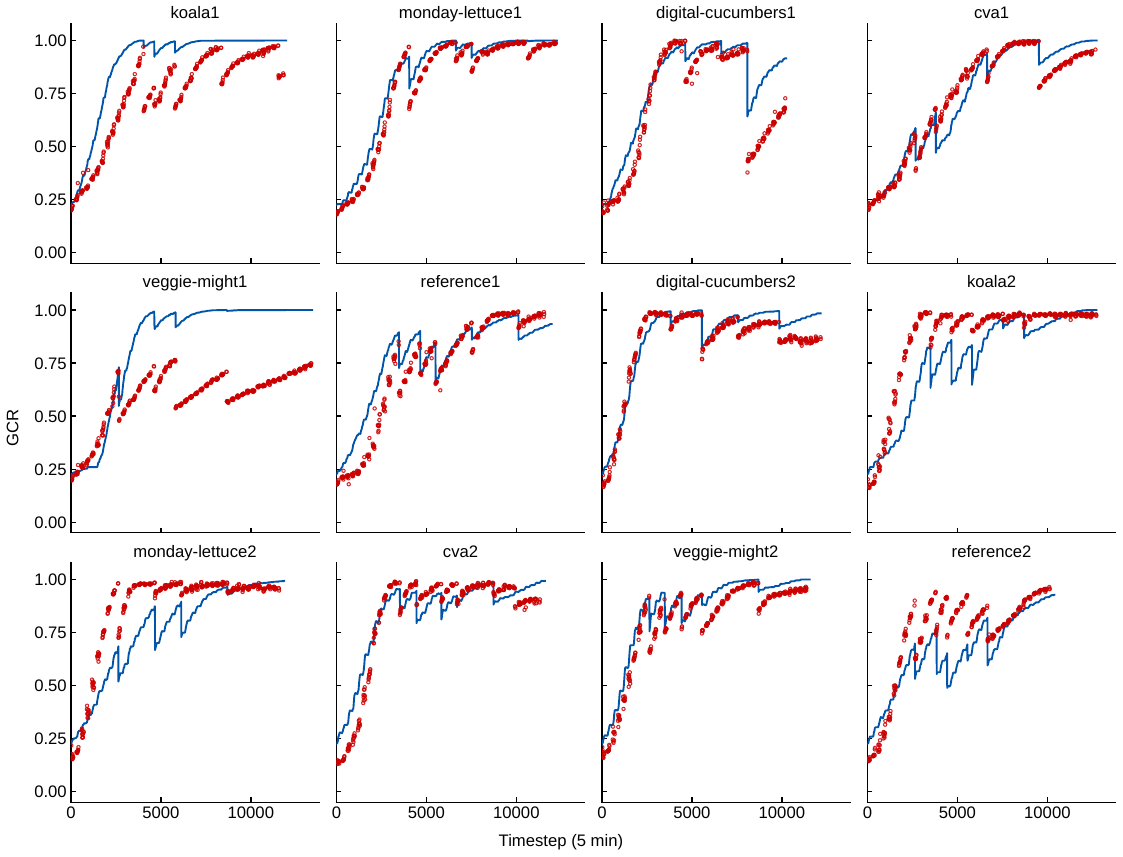}
        \caption{Test-strategy Ground coverage ratio (GCR) dynamics across cultivation strategies, simulated using the PBM-simulated shoot dry weight obtained from the six-fold leave-strategy-out cross-validation, with each strategy driven by the fold-specific parameters from the run in which it was held out. Simulated GCR (blue) is compared with image-derived GCR measurements (red). The per-strategy metrics are listed in Table~\ref{tab:F3}.}
        \label{fig:gcr_under_global_pbm}
    \end{figure}

    Fig.~\ref{fig:gcr_under_global_pbm} presents the comparison between GCR simulated from the PBM-simulated shoot dry weight of Experiment~1 and GCR derived from images across the 12 strategies. Because each input trajectory was generated using PBM parameters calibrated without it, this comparison tests the full pipeline on data that the pipeline had not seen during calibration. Averaged across strategies, the agreement was $\mathrm{R}^{2}=0.40$ with mean RRMSE of 16.8\%, substantially lower than that obtained when the representation was driven by measured shoot dry weight.
    
    This lower average was, however, driven mainly by a single strategy rather than by a uniform degradation across strategies. Close agreement between simulated and image-derived GCR was still shown by most strategies, and the main temporal trends were reproduced. The main exception was strategy \textit{veggie-might1}, where a large deviation resulted in a strongly negative $\mathrm{R}^{2}$ ($-3.56$) and the largest RRMSE (49.5\%). Excluding this strategy increased the average performance to $\mathrm{R}^{2}=0.76$, with a mean RRMSE of 13.8\% and strategy-wise $\mathrm{R}^{2}$ ranging from 0.38 to 0.95, which remained below the corresponding $\mathrm{R}^{2}=0.93$ achieved under measured shoot dry weight. Among the remaining strategies, \textit{koala1} showed the lowest agreement ($\mathrm{R}^{2}=0.38$, RRMSE 23.0\%).

\subsection{Experiment 4}

\begin{figure}[H]
        \centering
        \includegraphics[width=\linewidth]{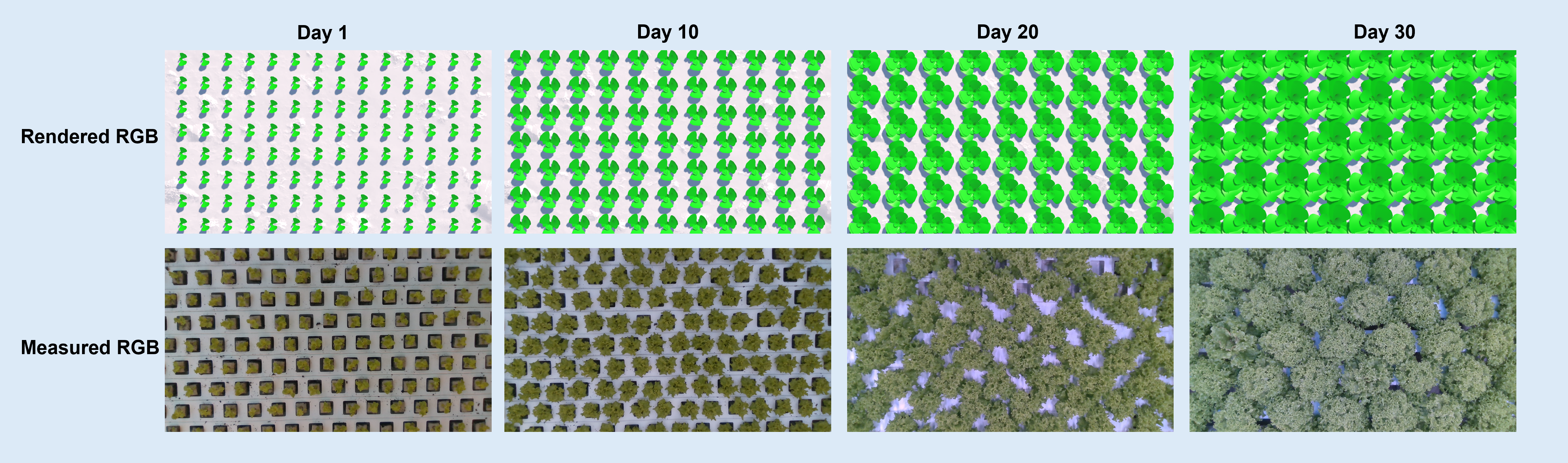}
        \caption{Visual comparison between rendered and measured RGB images at Days~1, 10, 20, and 30 (12:00) during cultivation. Measured images were obtained from the \textit{cva1} strategy, and rendered images were generated from canopy layout representations driven by measured shoot dry weight on the corresponding days. The rendering engine used an aspect ratio of 16:9 and an image resolution of $1920 \times 1080$, matching those of the measured images.}
        \label{fig:rendered_rgb_comparision}
    \end{figure}

    Fig.~\ref{fig:rendered_rgb_comparision} provides a visual comparison between the rendered and measured top-view RGB images at days~1, 10, 20 and 30. Overall, the major temporal trends were reproduced by the rendered images, including gradual increases in apparent plant size and canopy closure observed in the measurements. Minor discrepancies in spatial arrangement were attributable to the uniform-grid assumption inherent in the canopy layout representation.

\begin{figure}[H]
        \centering
        \includegraphics[width=\linewidth]{
            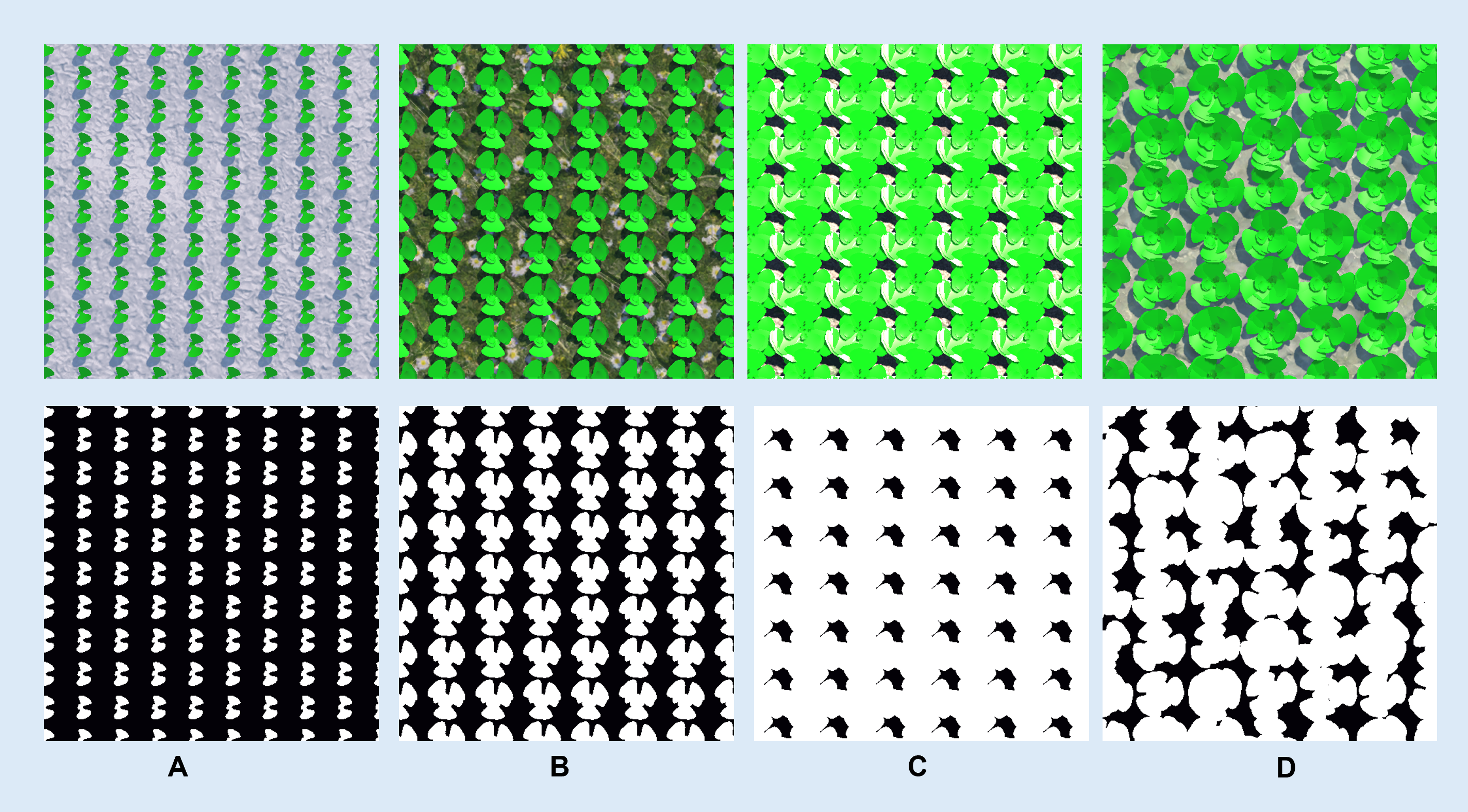
        }
        \caption{Visual illustration of rendered RGB images and corresponding
        segmentation masks under different rendering configurations using an
        identical canopy layout representation. Rendered RGB images (top row) and
        their corresponding segmentation masks (bottom row) were generated using
        a fixed square field of view ($L = W = 1$~m) and an image resolution of
        $512 \times 512$. All images were rendered from the same canopy layout representation
        simulated using the \textit{cva1} strategy at day 10. Columns (A)–(C) correspond to
        low, medium, and high lighting levels, respectively. Column (D) shows an
        additional configuration with medium lighting combined with randomised prefab
        scale and rotation. Different background textures were applied across
        configurations to illustrate visual variation in the rendered outputs.}
        \label{fig:rendered_different_configuration}
    \end{figure}

    As illustrated in Fig.~\ref{fig:rendered_different_configuration}, diverse visual outputs were generated by the rendering engine from an identical canopy layout representation. Variations in illumination and background texture (A–-C) altered the visual style without affecting the underlying geometry, which enables controllable appearance changes while keeping the underlying canopy structure and labels fixed. Conversely, the introduction of plant-level randomisation (D) modified the canopy geometry in a controlled manner, which was accurately reflected in the corresponding segmentation masks.

\begin{figure}[H]
        \centering
        \includegraphics[width=\linewidth]{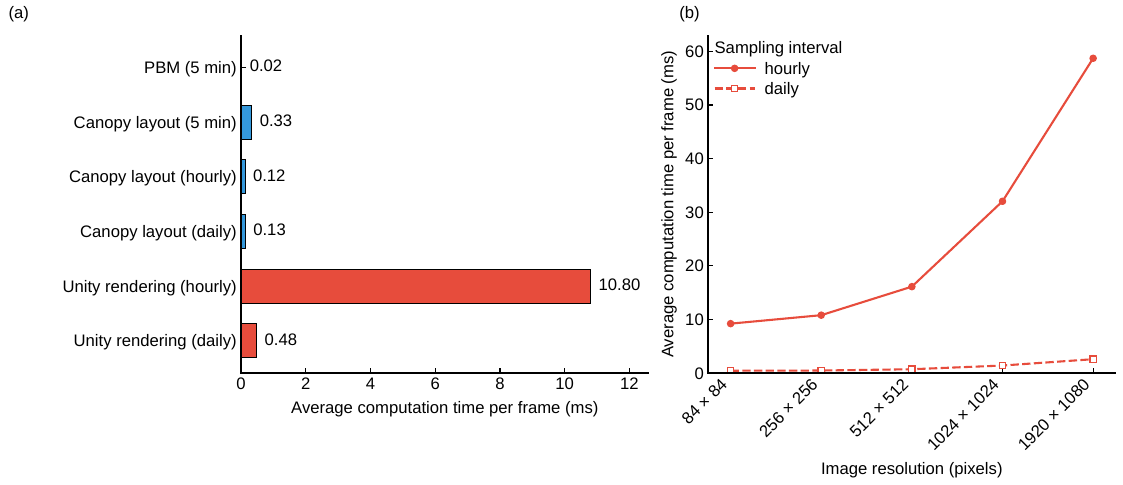}
        \caption{Comparison of computational efficiency across simulator components and rendering configurations. (A) Average computation time per frame measured under three execution scopes of the simulation pipeline: PBM-only execution, PBM combined with the canopy layout algorithm, and the full pipeline. The canopy layout representation was converted into a binary mask at an image resolution of $256 \times 256$~pixels, while the rendered images were generated with a square field of view (L=W=1~m) and an image resolution of $256 \times 256$~pixels. (B) Average computation time per frame of the full pipeline under different image resolutions and sampling intervals. All simulations were conducted over a 30-day cultivation period, with environmental conditions and plant spacing management of the \textit{cva1} strategy as inputs. Each reported value represents the mean of five independent runs; the standard deviation was below 0.005 in all cases and is therefore not shown.}
        \label{fig:simulator_computational_efficiency}
    \end{figure}

    The computational cost of the simulator and of the Unity rendering engine under different configurations is quantified in Fig.~\ref{fig:simulator_computational_efficiency}. As shown in Fig.~\ref{fig:simulator_computational_efficiency}(A), Unity rendering dominated the runtime, whereas the PBM update and canopy layout algorithm contributed substantially less per frame. Moreover, the cost of the canopy layout algorithm increased as the image sampling interval decreased, with 5~min updates requiring more computation than hourly updates. Under daily sampling, Unity rendering engine took approximately four times as long as the canopy layout algorithm. When the sampling interval was reduced from daily to hourly, the rendering time per frame increased by roughly one order of magnitude (about 22 times), while the PBM contribution remained comparatively small. The effect of image resolution is further shown in Fig.~\ref{fig:simulator_computational_efficiency}(B), for both sampling intervals, the average computation time per frame increased with resolution, and this increase was more pronounced under hourly sampling. At the highest tested resolution, the rendering time reached approximately 60~ms per frame under hourly sampling.

    \subsection{Experiment 5}

\begin{figure}[H]
        \centering
        \includegraphics[width=\linewidth]{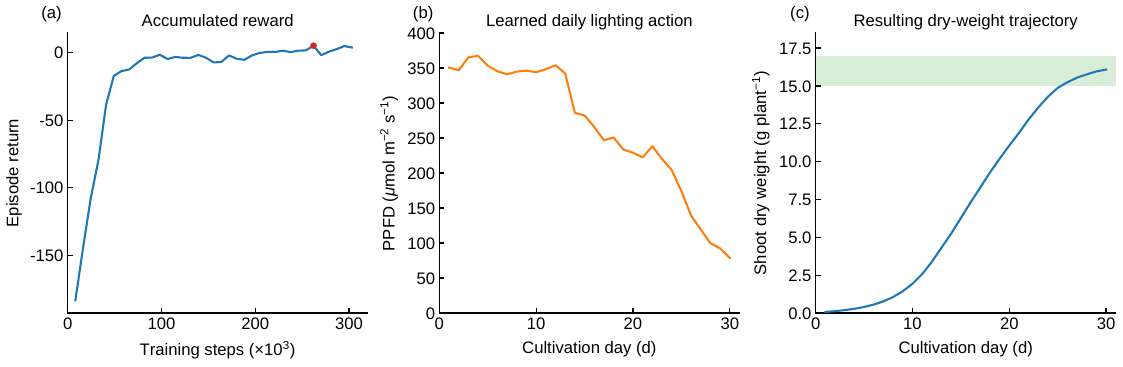}
        \caption{Vision-based reinforcement learning demonstration. (a) Accumulated episode return over training steps, recorded as the running mean of the episode return; the red marker indicates the maximum (max~$=5.09$). (b) Daily lighting action $a_{d}$ predicted by the trained policy. (c) Resulting per-day shoot dry weight trajectory from a single rollout. The shaded band is the harvest band $[15,17]~\mathrm{g}\cdot\mathrm{plant}^{-1}$.}
        \label{fig:rl_demonstration}
    \end{figure}

    It is demonstrated in Fig.~\ref{fig:rl_demonstration} that the lighting-control task was learned by the agent directly from images. Training progress is tracked by the accumulated reward, recorded as the running mean of the episode return (Fig.~\ref{fig:rl_demonstration}(a)), which rose steeply during the early steps and converged after around 100k steps, reaching a maximum episode return of about 5.09. The policy achieving the best accumulated reward was selected for examining the learned behaviour. Fig.~\ref{fig:rl_demonstration}(b) and (c) show, from a single rollout, the daily lighting action predicted by this policy and the resulting shoot dry weight accumulation across the 30-day cycle, respectively. The shoot dry weight reached 16.1~$\mathrm{g}\cdot\mathrm{plant}^{-1}$, within the harvest band.

    \section{Discussion}
    \label{sec:discussion}
    
    \subsection{Validation of the process-based model under dynamic plant density}

    Shoot dry weight was predicted reasonably well by the PBM under dynamic plant density in Experiment~1, as supported by the close agreement between the test and the training metrics ($\mathrm{R}^{2}=0.836\pm0.131$ vs $0.873\pm0.031$), which suggests that the calibrated parameters transferred to cultivation strategies not seen during calibration without substantial overfitting. This supports the use of the Van Henten PBM, despite its fixed-density origin, as the upstream growth model for dynamic-spacing scenarios within the scope of the cultivars and conditions examined here.

    The agreement nonetheless varied across strategies, and this variability was driven by a few difficult strategies rather than by a uniform loss of accuracy, as reflected in the larger spread of the test metrics relative to the train metrics. The clearest example was \textit{veggie-might1}: the fold containing this strategy showed the weakest test agreement (Table~\ref{tab:exp1_dw_cv}). This is also confirmed by the shoot dry weight trajectory in Fig.~\ref{fig:global_pbm_shoot_dry_weight}, where \textit{veggie-might1} deviated substantially from the measurements during the later growth stage. Because the shoot dry weight that drives the downstream canopy layout representation is provided by the PBM, the accuracy of this upstream prediction directly conditions the validity of the GCR reproduced by the full pipeline, as examined in the following subsections.

    Two limitations of this evaluation should be noted. First, although the $\mathrm{R}^{2}$ values indicate that the PBM captured the overall shoot dry weight dynamics well, the associated RRMSE values were comparatively large relative to those reported in other lettuce growth modelling studies \citep{SUN2025285}. Such biomass prediction errors may propagate through the downstream pipeline, indicating that the PBM still requires further improvement. Second, the analysis was restricted to a single cultivar; whether the calibrated PBM generalises equally well under dynamic plant density for other cultivars remains to be established.

    \subsection{Modelling choices for the PPA--shoot dry weight mapping}
 
    The comparison between the two-phase and one-phase regression models suggests that a stage-dependent strategy better captures the mapping $f_{\mathrm{PPA}}: w^{\ast} \mathrm{PPA}$. Although comparable overall cross-validation performance was achieved by several nonlinear one-phase models, the early growth stage ($w<w_{b}$) benefited most from the two-phase formulation. This pattern implies that the relation between PPA and shoot dry weight is not uniform across development, and that the breakpoint $w_{b}$ can be interpreted as a transition point at which the underlying relationship changes.

    In the context of this study, the value of this mapping lies less in standalone trait prediction than in providing an individual-plant structural parameter with which the canopy layout representation was simulated. From this perspective, improved accuracy at the early growth stage is particularly important. Errors in PPA estimation directly affect the projected canopy radius of individual plants. Under a high plant density, such errors can propagate across many plants when the canopy layout representation is constructed. The better early-stage fit therefore improves the validity of canopy layout representation, rather than merely improving an isolated regression task.

    The identified breakpoint also implies that the relationship between biomass accumulation and horizontal canopy expansion changes over the course of growth, which is plausible given the shifts in self-shading and rosette expansion that occur as lettuce grows \citep{geldhof_digital_2021,liu_far-red_2022}. In this sense, the two-phase formulation is not only an empirical improvement but also a compact way to reflect a development-dependent change in canopy structure.

    At the same time, two limitations of the present analysis should be acknowledged. First, the model comparison is based on average cross-validation metrics without statistical testing, so the numerical advantage of the two-phase model over the one-phase models should be interpreted with caution. Second, the proposed mapping was derived from a single lettuce cultivar, whereas previous work has shown that the association between morphological characteristics and shoot dry weight can differ across red and green cultivars \citep{kim_morphological_2022}; both the parameter values,and potentially, the functional form of $f_{\mathrm{PPA}}$ may therefore change for other cultivars. Future work should apply appropriate statistical tests to confirm whether the observed advantage is statistically significant, and should examine whether the two-phase structure remains valid when broader morphological diversity is included.

    \subsection{Validity of the canopy layout representation}

    The canopy layout representation serves as a structural proxy that links the crop state $w^{\ast}$ to visual observations. Because its construction relies on determining the projection radius $r=\sqrt{\mathrm{PPA}/\pi}$ via the empirical mapping $f_{\mathrm{PPA}}$, the validity of the resulting canopy layout representation $\mathcal{C}$ depends on whether it preserves canopy-scale structural information consistent with real measurements under dynamic conditions.

    This validity should be interpreted as functional consistency rather than exact geometric reconstruction. Qualitative comparisons of binary masks (Fig.~\ref{fig:layout_mask_visualization}) show that simulated and measured masks were visually similar at early growth stages, whereas more pronounced differences emerged as cultivation progressed: the simulated masks retained a regular grid-based organisation, while the measured masks showed more irregular uncovered gaps, irregular plant positions, and complex canopy boundaries. These discrepancies arise because within-canopy heterogeneity in plant size, shape and local position is neglected by the assumptions of circular canopy projection and uniform spacing. Such variability is instead partially delegated to the rendering stage, where spatial irregularity can be reflected by controlled plant-level randomisation (Fig.~\ref{fig:rendered_different_configuration}).

    When driven by measured shoot dry weight, the image-derived GCR was reproduced closely by the representation, with an overall $\mathrm{R}^{2}=0.84$ across all strategies which increased to $\mathrm{R}^{2}=0.93$ when \textit{veggie-might1} was excluded. This indicates that the dominant aggregate effects of crop state on canopy coverage are captured by the representation, even though the fine-scale irregularities seen in the masks are not reproduced by its simplified geometry. When instead driven by PBM-simulated shoot dry weight, the GCR agreement decreased to an overall $\mathrm{R}^{2}=0.40$ that increased to $\mathrm{R}^{2}=0.76$ when \textit{veggie-might1} was excluded (Fig.~\ref{fig:gcr_under_global_pbm}); the representation therefore remained a valid intermediate layer, and its downstream agreement depends on the quality of the input shoot dry weight. Notably, these evaluations were conducted on data independent of those used for calibration, in two senses: (1)~the parameters of $f_{\mathrm{PPA}}$ were estimated on Dataset~(i) and validated here on Dataset~(ii); and (2)~in the PBM-driven case, the shoot dry weight trajectories were produced by the leave-strategy-out cross-validation of Experiment~1, so the full pipeline was additionally evaluated on strategies unseen during calibration. Together, these demonstrate the generalisation of the proposed canopy layout representation, both on its own and within the full pipeline, to unseen dynamic environmental and plant-density conditions.

    The degradation under PBM-simulated shoot dry weight, and its variation across strategies, can be attributed to two stage-dependent error sources. Before canopy closure, the GCR is sensitive to shoot dry weight, so errors in the PBM-simulated shoot dry weight propagate into the GCR; this accounts for most of the gap between the measured- and PBM-driven cases. After canopy closure, however, the GCR saturates and becomes insensitive to shoot dry weight, so the remaining late-stage discrepancy reflects the idealised geometry of the representation rather than the input shoot dry weight: the uniform-grid, circular-overlap construction closes the canopy almost completely, whereas the measured canopy retains the irregular gaps seen in the mask comparison (Fig.~\ref{fig:layout_mask_visualization}), leaving the simulated GCR slightly above the measured one. Strategy \textit{veggie-might2} illustrates this interaction: although its PBM-simulated shoot dry weight was underestimated at the late growth stage (Fig.~\ref{fig:global_pbm_shoot_dry_weight}), the corresponding late-stage GCR was not underestimated but slightly overestimated under both measured and PBM-simulated inputs (Fig.~\ref{fig:GCR_under_measured_dry_weight} and Fig.~\ref{fig:gcr_under_global_pbm}), consistent with a saturation-dominated, dry-weight-independent error. An exception is \textit{koala2}, for which the shoot dry weight was predicted accurately yet the downstream GCR agreement still dropped markedly; this exceeds what either error source predicts and points instead to representation-level effects.

    The results also suggest the boundary conditions of the proposed representation. It is most suitable for highly controlled CEA systems with regular planting layouts and relatively homogeneous canopy development, such as plant factories and high-tech greenhouses. Its validity is limited where canopy development is strongly affected by unmodelled biological or management constraints, or where spatial irregularity plays a larger role. In addition, \textit{veggie-might1} not only failed consistently in the shoot dry weight and GCR simulations but also showed a measured GCR that remained below the level reached by most other strategies throughout cultivation, suggesting that canopy development in this strategy may have been restricted by biological or management factors not represented in the current formulation of the canopy layout representation.

    Another limitation may arise from the simplified imaging assumptions used to derive the simulated GCR. The measured GCR was obtained from real camera images subject to the characteristics of the physical imaging system, whereas the simulated GCR was derived from an idealised canopy layout representation under simplified viewing assumptions. Specifically, the GCR estimated from measured images may be influenced by image-formation factors such as perspective effects, camera intrinsic parameters, lens characteristics, and imperfect alignment between the camera view and the crop rows. None of these factors were modelled explicitly in this study. Consequently, this mismatch between the real imaging process and the idealised representation may contribute to both the visual (Fig.~\ref{fig:layout_mask_visualization}) and quantitative (Fig.~\ref{fig:gcr_under_global_pbm}) discrepancies discussed above, adding uncertainty to the validation beyond what the representation alone can account for. This uncertainty is nonetheless unlikely to alter substantially the overall finding that the proposed representation reproduces the dominant canopy-scale trends in GCR dynamics.

    Taken together, these results support the canopy layout representation as a useful intermediate layer for canopy-scale simulation, while also showing that its validity is conditional rather than universal. Future improvements should therefore proceed along two directions: first, by extending the canopy layout representation to account for environment-dependent structural heterogeneity at the plant level; and second, by designing validation experiments in which camera intrinsic parameters and misalignment between camera view and crop rows are characterised and controlled explicitly, enabling a cleaner separation between representation-level errors and discrepancies arising from image formation effects.

    \subsection{Implications and future work}
    Taken together, the results of this study suggest that LettuceVisSim constitutes a useful simulator for vision-based RL in CEA. Because biomass dynamics, canopy structure and image generation are represented as connected but modular components, an explicit link, the simulator preserves an explicit link between crop state and visual output while maintaining the computational efficiency required for scalable synthetic data generation.

    This modular perspective also has methodological implications. Because the PBM, the canopy layout algorithm and the rendering engine are separated but linked, the simulator is easier to interpret, extend and refine. The modular structure also facilitates error attribution: the results of Experiment~3, for example, show that reduced downstream GCR agreement can arise from inaccuracies in upstream shoot dry weight simulation rather than from the canopy layout representation alone. At the same time, the individual components can be replaced or improved without redesigning the full pipeline. A greenhouse crop model such as that of \citet{laatum_greenlight-gym_2024}, for instance, could be incorporated in place of the current PBM in order to simulate additional environmental variables simultaneously, which would extend the simulator's scope to support more complex CEA control problems.
    
    Future work should proceed along two main directions. The first concerns further improving simulator accuracy while preserving computational efficiency. This includes extending the shoot dry weight--PPA mapping and canopy layout representation to a broader range of cultivars and environmental conditions, introducing stronger links between environmental dynamics and plant-level structural heterogeneity, improving the treatment of camera alignment and image-formation effects in the validation pipeline, and enriching the prefab library to better represent cultivar-specific appearances and environment-dependent morphological responses.

    The second direction builds on the present validation to further explore vision-based RL for CEA. It was demonstrated in Experiment~5 that environment control can be learned from crop images by vision-based reinforcement learning. Future studies should extend this towards more realistic settings, including a broader range of control tasks and environmental conditions, comparison with baselines that rely on biomass measurements, and analyses of robustness, generalisation, and sim-to-real transfer.

    \section{Conclusion}
    \label{sec:conclude} 
    This study presents LettuceVisSim, a lettuce growth simulator that closes the data scarcity gap by generating labelled time-series crop images for vision-based RL in CEA. Validated across a broad range of dynamic conditions, the simulator's capability was established through five main findings. First, validation showed that the PBM predicted shoot dry weight under dynamic plant-density management with $\mathrm{R}^{2}=0.84$ on unseen cultivation strategies. Second, a two-phase piecewise cubic regression mapped shoot dry weight to PPA with $\mathrm{R}^{2}=0.94$. Third, the proposed canopy layout representation was shown to be valid under 12 previously unseen cultivation strategies, while its performance depends on the quality of the supplied shoot dry weight, $\mathrm{R}^{2}=0.84$ when driven by measured values vs $\mathrm{R}^{2}=0.40$ when driven by PBM-simulated values, increasing to $\mathrm{R}^{2}=0.76$ after excluding a single outlier strategy. Fourth, high computational efficiency was achieved by the Unity rendering engine, with a rendering time below 10~ms per frame at a low resolution. Fifth, it was demonstrated that a lighting-control policy can be learned and then operated by observing crop images without biomass estimation, providing the a proof of concept of vision-based reinforcement learning in CEA. Collectively, these results establish LettuceVisSim as a viable simulator for vision-based RL research in CEA.

    \section*{Declaration of competing interest}
    The financial support of this paper was provided by the China Scholarship Council. All authors declare that they have no known competing financial interests or personal relationships that could have appeared to influence the work reported in this paper.

    \section*{CRediT authorship contribution statement}
    \textbf{Ziye Zhu:} Conceptualisation, Data curation, Formal analysis, Investigation, Methodology, Software, Validation, Visualisation, Writing -- original draft, Writing -- review and editing.
    \textbf{Bert van 't Ooster:} Conceptualisation, Methodology, Writing -- review and editing.
    \textbf{Congcong Sun:} Methodology, Supervision, Writing -- review and editing.
    \textbf{Eldert van Henten:} Conceptualisation, Supervision, Writing -- review and editing.
    \textbf{Sjoerd Boersma:} Supervision, Writing -- review and editing.

    \section*{Acknowledgements}
    This work was supported by the China Scholarship Council (grant No. 202209110010). The authors thank the members of the Agricultural Biosystems Engineering (ABE) group at Wageningen University \& Research for their valuable discussions and constructive feedback throughout this work.

    \section*{Declaration of generative AI and AI-assisted technologies in the manuscript preparation process}
    During the preparation of this work the author(s) used ChatGPT in order to check for grammatical errors. After using this tool/service, the author(s) reviewed and edited the content as needed and take(s) full responsibility for the content of the published article.

    \section*{Data availability}
    The datasets analysed in this study are publicly available from the 3rd Autonomous Greenhouse Challenge \citep{hemming_3rd_2021,petropoulou_lettuce_2023}. A demonstration implementation of LettuceVisSim, including example configurations and scripts to run the simulator, is available at \url{https://github.com/ThomasZiyeZhu/LettuceVisSim.git}.

    \appendix
    \makeatletter
    \@addtoreset{figure}{section}
    \@addtoreset{table}{section}
    \makeatother
    \section{Process-based model equations}
    \label{appendix:pbm}
    
    This appendix summarises the process-based model (PBM) equations used in this study. Let $p$ denote the set of model parameters appearing in the PBM formulation. The governing equations are given below, and Table~\ref{tab:pbm_parameters} lists the definitions, values, and units of the individual parameters in $p$.

    \begin{align}
        \dot X_n &= c_a f_{\mathrm{phot}} - r_{\mathrm{gr}}X_s - f_{\mathrm{resp}} - \frac{1-c_b}{c_b}r_{\mathrm{gr}}X_s, \\
        \dot X_s &= r_{\mathrm{gr}}X_s,
    \end{align}
        where
    \begin{align}
        r_{\mathrm{gr}} &= c_{\mathrm{gr,max}}\frac{X_n}{c_gX_s+X_n}\,c_{Q10,\mathrm{gr}}^{(T-20)/10}, \\
        f_{\mathrm{resp}} &= \left[c_{\mathrm{resp,sht}}(1-c_t)X_s + c_{\mathrm{resp,rt}}c_tX_s\right]c_{Q10,\mathrm{resp}}^{(T-25)/10}, \\
        f_{\mathrm{phot}} &= \left(1-\exp\!\left[-c_K c_{\mathrm{lar}}(1-c_t)X_s\right]\right) f_{\mathrm{phot,max}}, \\
        f_{\mathrm{phot,max}} &= \frac{e I g_{\mathrm{CO_2}} c_w (C_{\mathrm{CO_2}}-G)}{eI + g_{\mathrm{CO_2}} c_w (C_{\mathrm{CO_2}}-G)}, \\
        G &= c_G c_{Q10,G}^{(T-20)/10}, \\
        e &= c_e \frac{C_{\mathrm{CO_2}}-G}{C_{\mathrm{CO_2}}+2G}, \\
        g_{\mathrm{CO_2}}^{-1} &= g_{\mathrm{bnd}}^{-1}+g_{\mathrm{stm}}^{-1}+g_{\mathrm{car}}^{-1}, \\
        g_{\mathrm{car}} &= -1.32\times10^{-5}T^2 + 5.94\times10^{-4}T - 2.64\times10^{-3}.
    \end{align}

    \begin{table}[H]
    \centering
    \caption{Model parameters used in the process-based lettuce growth model.}
    \label{tab:pbm_parameters}
    \begin{tabular}{lll}
    \hline
    Parameter & Definition & Unit \\
    \hline
    $c_a$ & CO$_{2}$ to CH$_{2}$O conversion factor & -- \\
    $c_b$ & Yield factor & -- \\
    $c_{\mathrm{gr,max}}$ & Saturation growth rate at 20$^\circ$C & s$^{-1}$ \\
    $c_g$ & Growth rate coefficient & -- \\
    $c_{Q10,\mathrm{gr}}$ & Temperature sensitivity of growth & -- \\
    $c_{\mathrm{resp,sht}}$ & Shoot maintenance respiration coefficient at 25$^\circ$C & s$^{-1}$ \\
    $c_{\mathrm{resp,rt}}$ & Root maintenance respiration coefficient at 25$^\circ$C & s$^{-1}$ \\
    $c_{Q10,\mathrm{resp}}$ & Temperature sensitivity of respiration & -- \\
    $c_t$ & Root-to-total dry mass ratio (hydroponic) & -- \\
    $c_K$ & Extinction coefficient & -- \\
    $c_{\mathrm{lar}}$ & Structural leaf area ratio & m$^2$ g$^{-1}$ \\
    $c_w$ & CO$_{2}$ density & g m$^{-3}$ \\
    $c_G$ & CO$_{2}$ compensation point at 20$^\circ$C & mL L$^{-1}$ \\
    $c_{Q10,G}$ & Temperature sensitivity of CO$_{2}$ compensation point & -- \\
    $c_e$ & Quantum use efficiency & g J$^{-1}$ \\
    $g_{\mathrm{bnd}}$ & Boundary layer conductance & m s$^{-1}$ \\
    $g_{\mathrm{stm}}$ & Stomatal conductance & m s$^{-1}$ \\
    \hline
    \end{tabular}
    \end{table}
    
    \section{DeepLabv3+ training and validation performance}
    \label{appendix:deeplabv3}

\begin{figure}[H]
            \centering
            \includegraphics[width=1\linewidth]{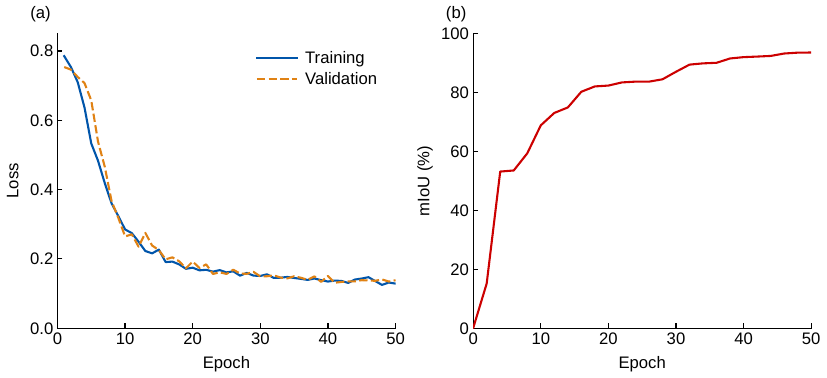}
            \caption{Training performance of DeepLabv3+. Left: training and validation loss over the training epochs. Right: validation mean intersection over union (mIoU) over the training epochs.}
            \label{fig:deeplabv3training}
        \end{figure}

    The performance of the segmentation model was quantified with the same mean intersection over union (mIoU) metric as used by \citet{petropoulou_lettuce_2023}, which is expressed as a percentage and is defined as

    \begin{equation}
        \mathrm{mIoU} = \frac{1}{2}\left(
        \frac{\left| M_{GT,\mathrm{lettuce}} \cap M_{\mathrm{lettuce}} \right|}{\left| M_{GT,\mathrm{lettuce}} \cup M_{\mathrm{lettuce}} \right|}
        + \frac{\left| M_{GT,\mathrm{background}} \cap M_{\mathrm{background}} \right|}{\left| M_{GT,\mathrm{background}} \cup M_{\mathrm{background}} \right|}
        \right) \times 100\%
        \label{eq:miou}
    \end{equation}

    \noindent in which $M$ denotes the mask of the lettuce class and of the background class, respectively, and $|\cdot|$ denotes the number of pixels in a mask. The mIoU of 93.5\% reported in Section~\ref{lab:data-selection} was obtained on the held-out test set. The training and validation loss curves and the validation mIoU over the training epochs are shown in Fig.~\ref{fig:deeplabv3training}.

    \section{Plant density and annotated image-based lettuce counts}
    \label{pd-and-plants-count}
\begin{figure}[H]
            \centering
            \includegraphics[width=1\linewidth]{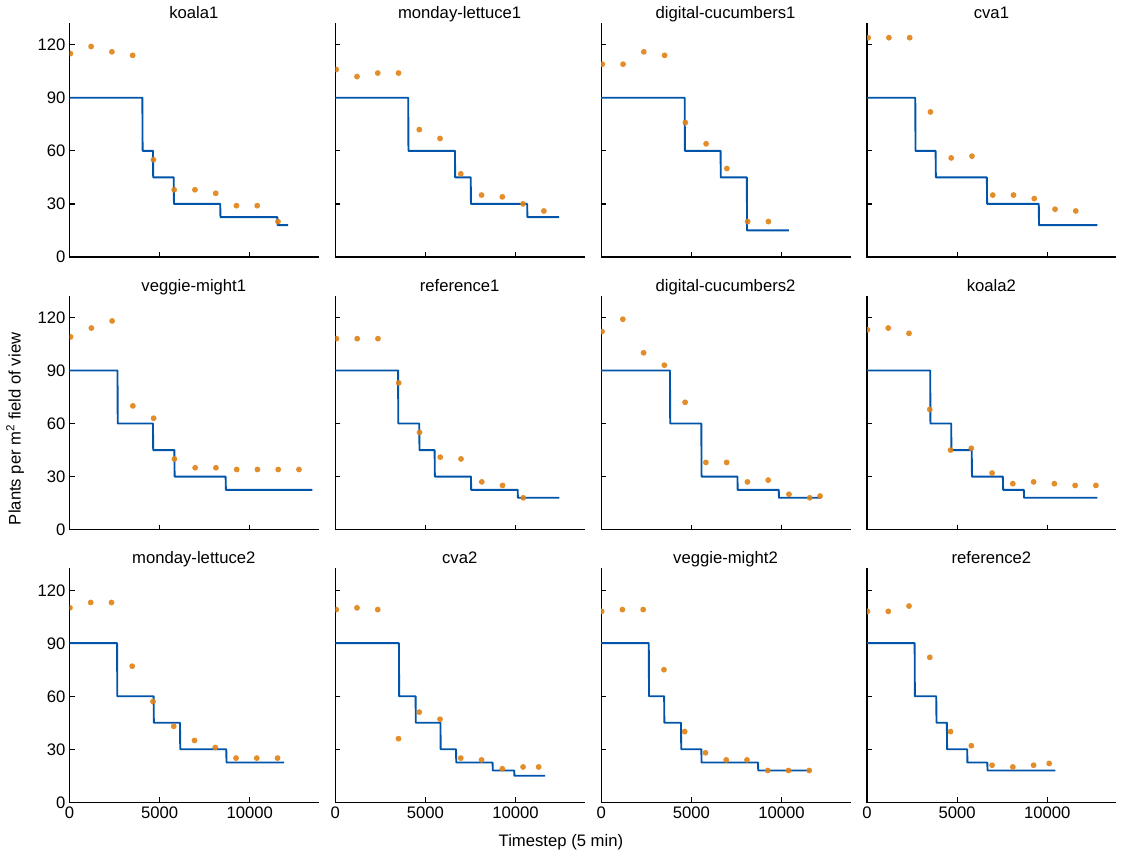}
            \caption{Temporal changes in plant density $pd$ and in the visible plant counts within the camera field of view, for each of the 12 cultivation strategies.}
            \label{fig:pd-and-plants-count}
        \end{figure}

    The relation between the recorded plant density and the number of lettuce plants visible within the camera field of view, as derived from the annotated images, is shown in Fig.~\ref{fig:pd-and-plants-count}

    \section{Evaluation metrics}
    \label{appendix:metrics}

    Throughout this study, predictive performance is quantified using the coefficient of determination ($\mathrm{R}^{2}$) and the relative root mean squared error (RRMSE). Let $o_i$ denote the $i$th observed value, $\hat{o}_i$ the corresponding predicted or simulated value, $\bar{o}$ the mean of all observed values, and $n$ the total number of observations. The metrics are defined as

    \begin{equation}
        \mathrm{R}^{2} = 1 - \frac{\displaystyle\sum_{i=1}^{n}(o_i - \hat{o}_i)^2}
                                   {\displaystyle\sum_{i=1}^{n}(o_i - \bar{o})^2},
        \label{eq:r2}
    \end{equation}

    \begin{equation}
        \mathrm{RMSE} = \sqrt{\frac{1}{n}\sum_{i=1}^{n}(o_i - \hat{o}_i)^2},
        \label{eq:rmse}
    \end{equation}

    \begin{equation}
        \mathrm{RRMSE} = \frac{\mathrm{RMSE}}{\bar{o}} \times 100\%.
        \label{eq:rrmse}
    \end{equation}

    The specific roles of $o_i$ and $\hat{o}_i$ differ per experiment; the precise variable definitions are given in the description of each experiment in Section~\ref{lab:experiments}.

    \section{Vision-based reinforcement network architecture and training details}
    \label{appendix:rl}

    This appendix details the network architecture and training configuration used in the vision-based RL demonstration.

    \paragraph{Network architecture.} An asymmetric actor--critic architecture was used. The actor encodes the $84\times84$ RGB image observation with a convolutional network into a 256-dimensional feature vector and therefore depends on the image alone; a fully connected head then outputs a Gaussian over a normalised action in $[0,1]$, rescaled to the physical range $[0,360]~\mu\mathrm{mol\ m}^{-2}\mathrm{\ s}^{-1}$. The critic uses the same image encoder but additionally concatenates the privileged true shoot dry weight (a scalar) during training, and outputs a scalar state-value estimate through a fully connected head. The implementation was based on Stable-Baselines3~\citep{raffin_stable-baselines3_2021}.

    \paragraph{Training setup.} The full set of PPO hyperparameters is listed in Table~\ref{tab:ppo_hyperparameters}.

    \begin{table}[H]
        \centering
        \caption{PPO training hyperparameters used in the vision-based RL demonstration.}
        \label{tab:ppo_hyperparameters}
        \begin{tabular}{lll}
            \toprule
            \textbf{Hyperparameter} & \textbf{Value} & \textbf{Description} \\
            \midrule
            Learning rate              & $3 \times 10^{-4}$  & Adam optimiser learning rate \\
            Discount factor            & 0.99                & Future reward discount \\
            GAE parameter              & 0.95                & Generalized advantage estimation \\
            Clip range                 & 0.2                 & PPO surrogate objective clipping \\
            Entropy coefficient        & 0.0                 & Exploration bonus weight \\
            Value function coefficient & 0.5                 & Critic loss scaling \\
            Gradient norm clipping     & 0.5                 & Maximum gradient norm \\
            Steps per environment      & 2048                & Rollout length per environment \\
            Mini-batch size            & 64                  & Batch size for gradient updates \\
            Number of epochs           & 10                  & Passes over collected data per update \\
            Total timesteps            & $3 \times 10^{5}$   & Training duration \\
            \bottomrule
        \end{tabular}
    \end{table}

    \section{Per-strategy evaluation metrics}\label{app:perstrategy}

    The per-strategy evaluation metrics that accompany Fig.~\ref{fig:global_pbm_shoot_dry_weight}, Fig.~\ref{fig:GCR_under_measured_dry_weight} and Fig.~\ref{fig:gcr_under_global_pbm} are listed in Table~\ref{tab:F1}, Table~\ref{tab:F2} and Table~\ref{tab:F3}, respectively. 

    \begin{table}[H]
    \centering
    \caption{Per-strategy agreement between the shoot dry weight simulated by the process-based model and the measured shoot dry weight, for the 12 test trajectories shown in Fig.~\ref{fig:global_pbm_shoot_dry_weight}. Each trajectory was generated with the fold-specific parameters from the cross-validation run in which that strategy was held out. $n$ is the number of destructive sampling occasions of that strategy.}
    \label{tab:F1}
    \begin{tabular}{lrrrr}
    \toprule
    Strategy & $R^{2}$ & RMSE (g plant$^{-1}$) & RRMSE (\%) & $n$ \\
    \midrule
    \textit{koala1} & 0.632 & 2.849 & 60.1 & 8 \\
    \textit{monday-lettuce1} & 0.991 & 0.400 & 9.8 & 8 \\
    \textit{digital-cucumbers1} & 0.728 & 1.130 & 53.6 & 7 \\
    \textit{cva1} & 0.985 & 0.616 & 13.7 & 8 \\
    \textit{veggie-might1} & $-$1.026 & 4.381 & 144.0 & 8 \\
    \textit{reference1} & 0.970 & 0.629 & 17.6 & 7 \\
    \textit{digital-cucumbers2} & 0.860 & 2.223 & 33.6 & 8 \\
    \textit{koala2} & 0.953 & 1.400 & 22.4 & 8 \\
    \textit{monday-lettuce2} & 0.746 & 2.141 & 43.6 & 8 \\
    \textit{cva2} & 0.952 & 1.328 & 21.4 & 8 \\
    \textit{veggie-might2} & 0.957 & 1.422 & 19.6 & 8 \\
    \textit{reference2} & 0.858 & 1.737 & 36.0 & 7 \\
    \bottomrule
    \end{tabular}
    \end{table}

    \begin{table}[H]
    \centering
    \caption{Per-strategy agreement between the ground coverage ratio simulated with measured shoot dry weight and the daily averaged image-derived ground coverage ratio, for the 12 cultivation strategies shown in Fig.~\ref{fig:GCR_under_measured_dry_weight}. $n$ is the number of measurement days of that strategy.}
    \label{tab:F2}
    \begin{tabular}{lrrrr}
    \toprule
    Strategy & $R^{2}$ & RMSE & RRMSE (\%) & $n$ \\
    \midrule
    \textit{koala1} & 0.89 & 0.088 & 12.2 & 8 \\
    \textit{monday-lettuce1} & 0.99 & 0.032 & 4.2 & 8 \\
    \textit{digital-cucumbers1} & 0.88 & 0.095 & 14.9 & 7 \\
    \textit{cva1} & 0.95 & 0.065 & 9.2 & 8 \\
    \textit{veggie-might1} & $-$0.25 & 0.201 & 36.3 & 8 \\
    \textit{reference1} & 0.93 & 0.076 & 12.2 & 6 \\
    \textit{digital-cucumbers2} & 0.91 & 0.079 & 10.0 & 8 \\
    \textit{koala2} & 0.96 & 0.061 & 7.5 & 8 \\
    \textit{monday-lettuce2} & 0.91 & 0.090 & 11.5 & 7 \\
    \textit{cva2} & 0.98 & 0.052 & 7.0 & 7 \\
    \textit{veggie-might2} & 0.98 & 0.043 & 5.8 & 7 \\
    \textit{reference2} & 0.94 & 0.063 & 9.2 & 6 \\
    \bottomrule
    \end{tabular}
    \end{table}

    \begin{table}[H]
    \centering
    \caption{Per-strategy agreement between the ground coverage ratio simulated with the shoot dry weight of the process-based model and the image-derived ground coverage ratio, for the 12 cultivation strategies shown in Fig.~\ref{fig:gcr_under_global_pbm}. Each strategy was driven by the fold-specific parameters from the cross-validation run in which it was held out. $n$ is the number of individual images of that strategy.}
    \label{tab:F3}
    \begin{tabular}{lrrrr}
    \toprule
    Strategy & $R^{2}$ & RMSE & RRMSE (\%) & $n$ \\
    \midrule
    \textit{koala1} & 0.38 & 0.170 & 23.0 & 412 \\
    \textit{monday-lettuce1} & 0.95 & 0.057 & 7.3 & 428 \\
    \textit{digital-cucumbers1} & 0.74 & 0.137 & 19.6 & 360 \\
    \textit{cva1} & 0.93 & 0.064 & 8.8 & 440 \\
    \textit{veggie-might1} & $-$3.56 & 0.292 & 49.5 & 469 \\
    \textit{reference1} & 0.85 & 0.103 & 14.4 & 403 \\
    \textit{digital-cucumbers2} & 0.91 & 0.060 & 7.1 & 421 \\
    \textit{koala2} & 0.55 & 0.149 & 17.3 & 443 \\
    \textit{monday-lettuce2} & 0.54 & 0.153 & 17.8 & 403 \\
    \textit{cva2} & 0.91 & 0.078 & 9.5 & 393 \\
    \textit{veggie-might2} & 0.91 & 0.067 & 8.4 & 400 \\
    \textit{reference2} & 0.64 & 0.135 & 18.5 & 353 \\
    \bottomrule
    \end{tabular}
    \end{table}
    
    \clearpage
    \bibliography{references}


\end{document}